\documentclass[letterpaper]{article} 
\usepackage{aaai2026}  
\nocopyright
\usepackage{times}  
\usepackage{helvet}  
\usepackage{courier}  
\usepackage[hyphens]{url}  
\usepackage{graphicx} 
\usepackage{natbib}  
\usepackage{caption} 
\usepackage{placeins}
\usepackage{amsmath}
\usepackage{amssymb}
\usepackage{booktabs}
\usepackage{listings}

\newcommand{\cofirst}{\textsuperscript{\ensuremath{\dagger}}} \newcommand{\corresponding}{\textsuperscript{*}}

\title{MapSatisfyBench: Benchmarking Satisfaction-Aware Map Agents through Behavior-Grounded Implicit Decision Factors}

\author{ Lubin Bai\textsuperscript{\rm 2}\cofirst, Mengyu Cao\textsuperscript{\rm 1}\cofirst, Sixue Wang\textsuperscript{\rm 1}, Zhongwei Wan\textsuperscript{\rm 1},\\ Yue Pan\textsuperscript{\rm 1}, Jiale Hou\textsuperscript{\rm 1}, Xiang Li\textsuperscript{\rm 1}\corresponding, Xiuyuan Zhang\textsuperscript{\rm 2}\corresponding } \affiliations{ \textsuperscript{\rm 1}AMAP, Alibaba Group, Beijing, China\\ \textsuperscript{\rm 2}College of Urban and Environmental Sciences, Peking University, Beijing, China\\ \ensuremath{\dagger} Equal contribution. * Corresponding authors. }

\begin{document}

\maketitle

\begin{abstract}
Large language model agents are increasingly integrated into map services. Since map services are embedded in everyday-life scenarios rather than professional task settings, users often express their needs informally, resulting in underspecified queries with many unspoken needs, namely, implicit decision factors that are critical for user satisfaction. Although clarification is an effective way to mitigate this issue, it increases user burden in daily interaction, and a capable agent should first proactively recover such factors from available information sources. However, evaluating this ability is challenging. The first challenge is to determine which implicit decision factors are suitable for evaluation. A factor is evaluable only if it affects user acceptance and can be recovered from information available to the agent before it responds. Second, user satisfaction cannot be reliably represented by a single reference answer, requiring a benchmark that converts satisfaction-relevant factors into objective and quantifiable evaluation targets. To address these challenges, we propose a restore-identify-filter framework that reconstructs complete user needs from behavior-chain evidence, identifies implicit decision factors, and retains only those supported by pre-query evidence. Building on this methodology, we construct MapSatisfyBench from large-scale, real-world anonymized user data and annotate ground truth from five dimensions and enables full-chain evaluation of satisfaction-aware map agents. Experiments show that current agents generally perform well on explicit task completion, but remain limited in satisfying implicit decision factors and proactively acquiring the evidence needed for satisfaction-aware decisions. These findings establish MapSatisfyBench as a benchmark for shifting map-agent evaluation from task completion toward satisfaction-aware spatial decision making.
\end{abstract}

\section{1  Introduction}

\begin{figure*}[t]
  \centering
  \includegraphics[width=0.92\textwidth]{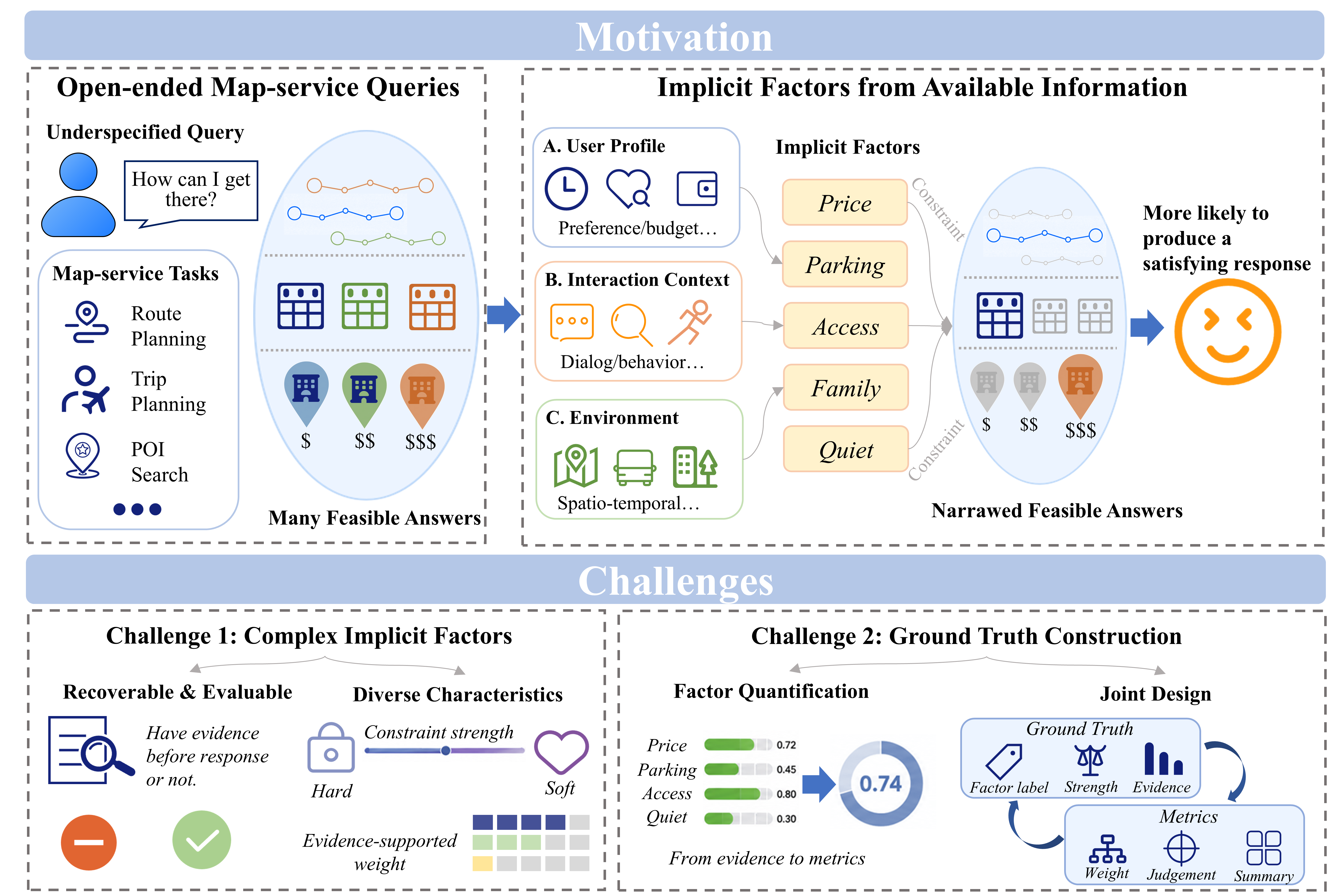}
  \caption{Motivation of MapSatisfyBench. Map-service queries often define multiple feasible responses, and satisfaction depends on whether the agent recovers behavior-supported implicit decision factors.}
  \label{fig:intro}
\end{figure*}

Large language model (LLM) agents are increasingly deployed as map-service assistants that translate natural language requests into executable actions, such as location search, route navigation, and trip planning ~\citep{xie2024travelplanner, hu2025amap}. By integrating natural language interaction, contextual signals, and tool-based execution, these agents provide more convenient and flexible support for everyday spatial decisions. However, since map services are embedded in everyday-life scenarios rather than professional task settings, real-world user queries are rarely fully specified ~\citep{kamvar2006wireless, church2009mobileintent}. Users often issue short and incomplete queries, while still expecting the agent to understand their needs even when those needs are not explicitly stated. This challenge is further amplified by the open-ended nature of many map-service tasks. Unlike fact-seeking questions that usually have a unique correct answer, many daily map queries are inherently multi-solution, i.e., their surface text defines a space of possible answers rather than a single ground-truth response ~\citep{vansteenwegen2011orienteering,purves2018geographic,delling2017customizable}. Consequently, the key mission for a smart map agent is not merely to produce a feasible response, but to select, among multiple feasible responses, the one most likely to satisfy the user.

Such satisfaction is closely tied to implicit decision factors that are not expressed in the query but are critical to whether the resulting decision is acceptable to the user ~\citep{pu2012evaluating,adomavicius2005toward}. For example, when a user asks "How can I get there?" after searching for a hospital near a railway station, both driving and public transit may be semantically valid responses; however, if the user has just arrived by train and has no car available, a transit-oriented route is more likely to be acceptable. Although an agent could resolve such uncertainty by asking the user for clarification, repeatedly doing so will increase interaction burden and reduce usability ~\citep{zou2023asking, zamani2020clarification}. In many cases, the missing factors can be recovered from available information source, like user profiles and interaction history ~\citep{church2009mobileintent,villegas2018contextaware}. A capable map agent should therefore proactively exploit these information sources and reserve clarification questions for cases that cannot be reliably resolved otherwise ~\citep{maes1994agents, horvitz1999mixedinitiative}. This motivates the study to design a benchmark for assessing whether map agents can identify information gaps, proactively acquire relevant evidence from available sources, and incorporate the recovered implicit decision factors into satisfaction-aware responses.

Recent benchmarks have substantially advanced the evaluation of LLM-based map agents from multiple perspectives, including planning, tool use, and information synthesis ~\citep{xie2024travelplanner,chaudhuri2025tripcraft,cheng2025travelbench,song2026mobilitybench,he2025localsearchbench,lbsintentbench2025}. These efforts provide important signals for improving map agents, but they do not directly measure whether an agent can make decisions that are likely to satisfy the user under underspecified map-service interactions. Building such a benchmark is meaningful but challenging in two respects. First, implicit decision factors are complex. They are the key conditions that narrow the feasible solution space and make one response more acceptable than other semantically valid alternatives, yet they are very complex. On the one hand, not every factor revealed by user behavior is suitable for evaluation, because some of them are not recoverable and evaluable. And a benchmark should retain only those factors that an agent could reasonably recover from information available before it responds (like historical behavior and spatio-temporal environment). On the other hand, These factors also have different characteristics. For example, some are hard constraints whose violation has a stronger negative impact on response acceptability, whereas others are soft preferences that only modestly shift the relative acceptability of feasible options. Second, ground truth construction is non-trivial because satisfaction cannot be reduced to a single reference answer or a direct satisfied/unsatisfied label. A satisfaction-oriented benchmark should instead convert the factors that affect acceptance into quantifiable evaluation references, including whether each implicit factor is satisfied and how strongly it should influence the final score. This requires the joint design of ground-truth annotations and evaluation metrics, so that the benchmark can comprehensively and accurately assess whether an agent produces responses that are likely to satisfy the user.

To address these challenges, we propose a restore-identify-filter framework based on behavior-chain evidence and  construct MapSatisfyBench. Behavior-chain evidence connects the pre-query information, the spatio-temporal environment, and the user's subsequent actions, providing an objective basis for reconstructing what the user was trying to accomplish. Based on this evidence, the restoration step recovers the complete need behind the interaction; the identification step compares this reconstructed need with the surface query to expose unspoken decision factors; and the filtering step retains only the factors supported by pre-response evidence and therefore evaluable for an agent. The retained factors are further annotated with their constraint type and evidence-supported weight, allowing MapSatisfyBench to quantify not only whether an agent satisfies the explicit task, but also whether it satisfies the implicit factors that affect accepted-response probability. On this basis, MapSatisfyBench provides a behavior-grounded benchmark for satisfaction-aware decision making in map services. Unlike task-completion-oriented benchmarks that primarily assess whether an agent can plan, retrieve, or execute a valid map-service task, MapSatisfyBench evaluates whether an agent can align open-ended map-service decisions with behavior-supported factors that determine whether a user is likely to accept the response.

MapSatisfyBench aims to promote the evaluation of map agents from "whether the task is finished" to "whether the decision satisfies the user." In our evaluation of 12 LLM-based agents, current systems generally achieve high scores on explicit intent completion and factual faithfulness, yet show clear weaknesses in implicit-need satisfaction and tool selection. These results indicate that map agents can often understand the stated request, but still struggle to identify the behavior-supported decision factors that make a response feel acceptable and useful to the user. Our contributions are threefold. First, we propose a methodology for converting subjective satisfaction into behavior-supported implicit decision factors, enabling objective evaluation without direct satisfaction labels. Second, we construct MapSatisfyBench, a satisfaction-aware benchmark covering diverse realistic map-service scenarios, together with a full-chain evaluation protocol that jointly assesses explicit task completion, implicit-need satisfaction, tool selection, and so on. Third, our experiments reveal a persistent gap between surface task completion and satisfaction-aware decision making, offering practical diagnostic insights for developing map agents that can better support real user decisions.

\section{2  Related Work}

\paragraph{Agent Benchmarks for Map Services.} The increasing deployment of LLM agents in map services has motivated a growing body of benchmarks for evaluating agent capabilities under geographic constraints and real-world service requirements. Early efforts use travel planning as a structured proxy for spatial decision making. TravelPlanner~\citep{xie2024travelplanner} introduced a first realistic sandboxes for evaluating multi-constraint itinerary generation with external tools and travel records, while subsequent benchmarks such as TripCraft~\citep{chaudhuri2025tripcraft}, TravelBench~\citep{cheng2025travelbench}, and VitaBench~\citep{he2025vitabench} extend evaluation toward more realistic service-oriented settings like unsolvable requests and cross-scenario life-service tasks. Collectively, these benchmarks mark a shift from static itinerary generation to interactive, tool-grounded service completion. More recent works evaluate agents in direct map-service scenarios. MobilityBench~\citep{song2026mobilitybench} evaluates route-planning agents on anonymized Amap queries through a deterministic API-replay sandbox, whereas LocalSearchBench~\citep{he2025localsearchbench} targets local-service search over large-scale merchant databases and real user requests requiring multi-hop reasoning. LBS-IntentBench~\citep{lbsintentbench2025} moves map agent evaluation beyond explicit instruction execution by focusing on implicit intent inference, which is partially aligned with our motivation. However, it primarily evaluates intent recovery. In contrast, MapSatisfyBench focuses on whether agents can make satisfactory and executable decisions when ambiguity remains, by reconstructing behavior-supported implicit decision factors from user behavior chains and evaluating satisfaction-aware responses.

\paragraph{Satisfaction-aware Agent Benchmarks.} User satisfaction has been studied from early dialogue-system evaluation to recent user-centric agent benchmarks. The classical PARADISE framework~\citep{walker1997paradise} relates satisfaction to task success and interaction cost, and later task-oriented dialogue work, including USS~\citep{sun2021uss}, SG-USM~\citep{feng2023schema}, SPUR~\citep{lin2024interpretable}, and CAUSE~\citep{abolghasemi2024cause}, further models satisfaction from dialogue trajectories, task-attribute fulfillment, interpretable rubrics, or counterfactual dissatisfaction cases. In general LLM evaluation, satisfaction is often approximated through human preference or LLM-as-judge protocols, as in InstructGPT~\citep{ouyang2022training}, WebGPT~\citep{nakano2021webgpt}, MT-Bench and Chatbot Arena~\citep{zheng2023judging}, AlpacaFarm~\citep{dubois2023alpacafarm}, Arena-Hard~\citep{li2024arena}, and the user-reported-scenario benchmark URS~\citep{chen2024urs}. More recent work moves toward interaction and personalization: UserBench~\citep{qian2025userbench} evaluates preference discovery under underspecified goals, AURA~\citep{kim2025aura} analyzes user satisfaction across interactive planning stages, CollabLLM~\citep{wu2025collabllm} optimizes multi-turn collaboration toward long-term user benefit, and personalization benchmarks such as PersonalLLM~\citep{zollo2025personalllm}, PrefEval~\citep{zhao2025prefeval}, and PersonaLens~\citep{zhao2025personalens} evaluate adaptation to individual preferences or user profiles. These studies show a clear shift from exact-match correctness to user-centered evaluation, but most treat satisfaction as a response-level preference, predicted label, or general interaction score. MapSatisfyBench instead operationalizes satisfaction in map-service decisions through behavior-chain evidence: it reconstructs recoverable implicit decision factors, distinguishes hard constraints from soft preferences, and evaluates whether a tool-grounded spatial response increases accepted-response probability.

\section{3  Method}

\begin{figure*}[t]
  \centering
  \includegraphics[width=0.98\textwidth]{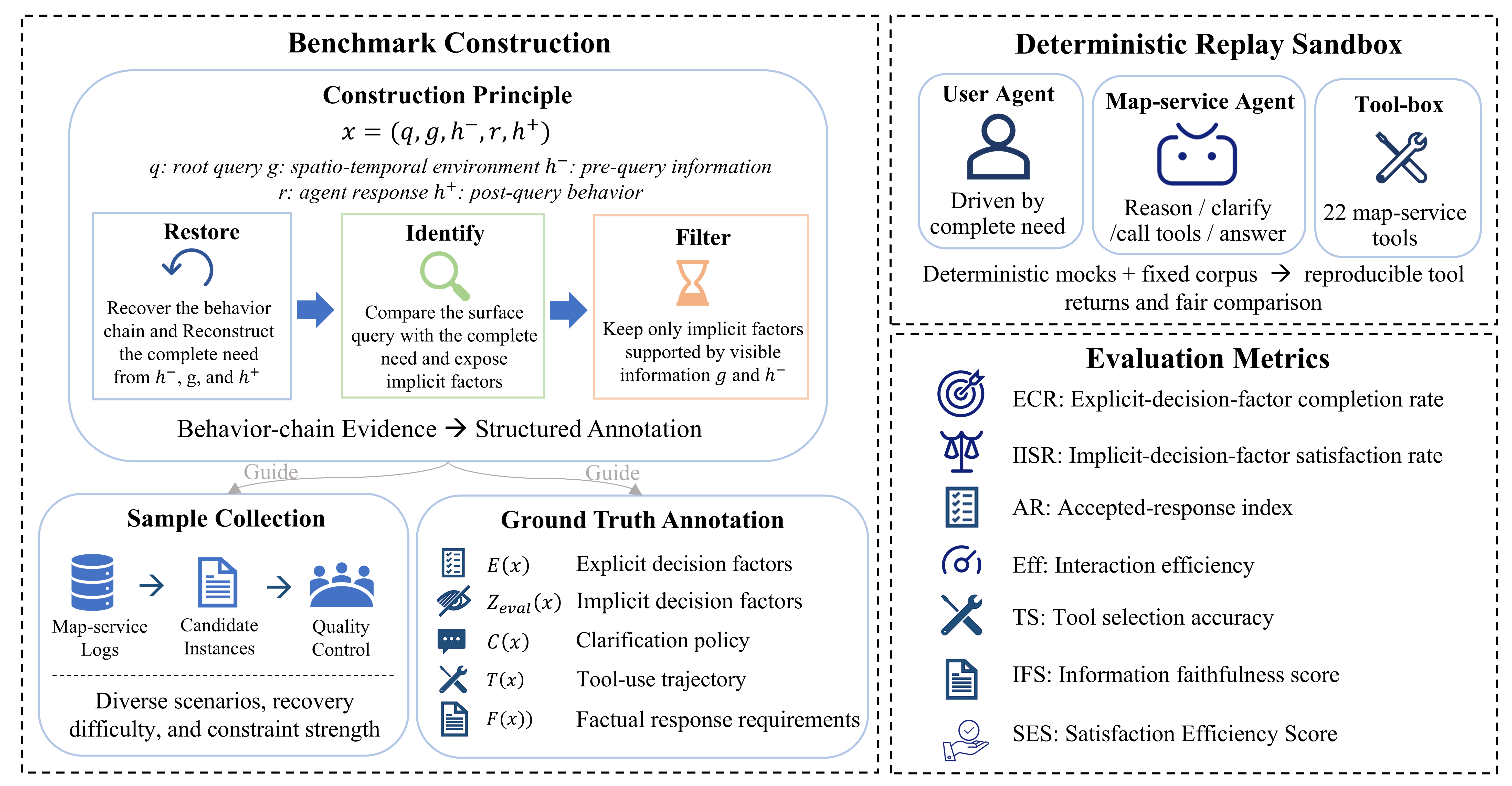}
  \caption{Overview of MapSatisfyBench. Benchmark construction follows the restore-identify-filter principle, the deterministic replay sandbox simulates tool-grounded interaction, and the evaluation metrics cover diverse aspects.}
  \label{fig:method}
\end{figure*}

\subsection{3.1 Benchmark construction}

\subsubsection{Construction Principle}

This section introduces the core logic of benchmark construction, which serves as the foundation for sample collection and ground truth construction. For each user instance, we formalize the interaction as:
\[
x = (q, g, h^{-}, r, h^{+}),
\]
where \(q\) is the root query, \(g\) denotes the spatio-temporal environment, \(h^{-}=(c,p)\) denotes the pre-query information available before the agent response, including interaction context \(c\) and historical profile \(p\), \(r\) is the agent response, and \(h^{+}\) denotes post-query user behavior after the agent responds. At inference time, the agent gets the query \(q\), produces a response \(r\) that narrows the feasible solution space by satisfying implicit needs inferred from the visible information \(g\) and \(h^{-}\), and ensures the post-query behavior \(h^{+}\) is not exposed to the agent.

MapSatisfyBench is constructed based on the principle of linking implicit needs with user satisfaction. We aim to identify all recoverable implicit constraints or preferences that may affect the user's final decision. These recovered needs must then be converted into structured ground truth and metrics for quantitative evaluation. To this end, we design a behavior-chain-based benchmark construction strategy with three steps: restore, identify, and filter.

The restoration step aims to recover the complete need behind a query. Specifically, we combine pre-query information \(h^{-}\), the spatio-temporal environment \(g\), and post-query behavior \(h^{+}\) to recover the user's behavior chain. Here, \(h^{-}\) is used to capture the contextual and historical signals available before the agent responds, \(g\) is used to anchor the request in the concrete time and location, and \(h^{+}\) is used to reveal the user's subsequent behavioral anchor and how the interaction actually unfolded. Together, these signals allow us to reconstruct the complete need behind the current interaction. The identification step aims to expose the gap between what the user explicitly states and what a satisfactory response must satisfy. we compare the surface query \(q\) with the reconstructed complete need to identify implicit needs that are not explicitly stated but can narrow the feasible solution space, such as transport feasibility, temporal constraints, or preference conditions. Not all identified needs are suitable for evaluation, because some cannot be inferred from information visible to the agent. Therefore, in the filtering step, we keep only the implicit needs whose evidence can be found in \(g\) and \(h^{-}\), which are the information sources available at decision time, and filter out factors that cannot be supported before the agent responds. The retained needs form an evaluable set of implicit decision factors, each of which is further labeled as a hard constraint or a soft preference. Through this process, MapSatisfyBench turns behavior-chain evidence into structured ground truth that can assess whether an agent's response satisfies both explicit task requirements and the implicit factors that affect user satisfaction. More details are provided in the Appendix A.1.
\subsubsection{Sample Collection.}
We collect benchmark instances by applying the restore-identify-filter principle to large-scale anonymized map-service logs. Given a candidate instance \(x=(q,g,h^{-},r,h^{+})\), we first restore the behavior chain around the root query and check whether the pre-query context, spatio-temporal environment, and post-query behavior together are sufficient enough to indicate a coherent map-service task. We then identify whether the surface query omits decision factors that would narrow the feasible solution space, which yields a candidate implicit-need set \(\mathcal{Z}(x)\). Finally, we filter this set by retaining only factors that are supported by information available before the agent responds, i.e., evidence in \(g\) and \(h^{-}\). A candidate is selected only when it satisfies the explicit task constraints, has a valid behavioral anchor in \(h^{+}\), and contains at least one retained implicit factor. This procedure keeps the benchmark grounded in naturally occurring user demands while ensuring that each instance contains implicit needs that are recoverable, evaluable, and relevant to user acceptance. The resulting sample collection covers diverse map-service scenarios and provides variation in implicit-need source, recovery difficulty, and constraint type, supporting both realistic coverage and discriminative evaluation of map agents.
\subsubsection{Ground Truth Annotation}
Ground truth Annotation aims to convert the factors that affect user satisfaction into quantifiable evaluation references. For each selected instance, MapSatisfyBench does not define a single gold answer. Instead, it builds a structured decision reference
\[
G(x)=\left(E(x), \mathcal{Z}_{\mathrm{eval}}(x), C(x), T(x), F(x)\right),
\]
where \(E(x)\) denotes explicit decision factors, \(\mathcal{Z}_{\mathrm{eval}}(x)\) denotes evaluable implicit decision factors, \(C(x)\) is the clarification policy, \(T(x)\) is the expected tool-use trajectory, and \(F(x)\) specifies factual response requirements. Separating explicit factors from implicit ones is necessary because a response may satisfy the literal request while still being unlikely to be accepted by the user. Explicit factors are extracted from the surface query to define the basic validity boundary of the task and provide the reference for evaluating whether the agent understands the stated request. On the other hand,  implicit factors are obtained from the restore-identify-filter procedure. They capture unspoken decision factors that are not stated in \(q\), but are supported by pre-response evidence and may change the probability that the user accepts the response. Thus, explicit factors define what any feasible answer must satisfy, whereas implicit factors define what a satisfaction-aware answer should additionally consider.

For each retained implicit factor \(z_i \in \mathcal{Z}_{\mathrm{eval}}(x)\), we further trace its evidence back to the information available at decision time, including the pre-query information \(h^{-}\) and the spatio-temporal environment \(g\). Based on this evidence, we annotate the ground truth with a scoring rubric, constraint type, and evidence-supported weight. Each implicit factor can be annotated as:
\[
z_i=(\rho_i, \tau_i, s_i),
\]
where \(\rho_i\) is the rubric instruction for evaluating whether the response satisfies the factor, \(\tau_i \in \{\mathrm{hard}, \mathrm{soft}\}\) indicates the constraint type, i.e., whether the factor is a necessary constraint or a graded preference, and \(s_i\) is its evidence-supported weight.

Hard constraints are factors whose violation has a relatively strong negative effect on response executability or acceptability, such as an unavailable transport mode or unresolved destination ambiguity. Soft preferences have a weaker, incremental effect on satisfaction and should therefore be weighted rather than treated as binary failures. Evidence-supported weight is a factor-level weight that reflects the expected influence of \(z_i\) on the accepted-response probability, and it is computed from both historical preference strength and current-session support:
\[
w_i = \mathrm{userpref}(z_i) \cdot \mathrm{CurrentNeed}(z_i).
\]
Here, \(\mathrm{userpref}(z_i)\) summarizes the support from long-term and recent user profile, while \(\mathrm{CurrentNeed}(z_i)\) measures whether this factor is active in the current interaction, considering relevance, continuity, directionality, and possible conflicts in the session. This design allows the benchmark to distinguish stable personal tendencies from temporary situational needs, and to assign higher weights to factors that are jointly supported by historical behavior and the current behavior chain.  More details are provided in the Appendix A.2.

In addition to explicit and implicit decision factors, each instance includes three execution-oriented annotations. The clarification policy \(C(x)\) specifies whether the agent should ask for additional information and the expected number of useful clarification turns, especially when an implicit factor is important but weakly supported. The tool-use trajectory \(T(x)\) specifies the expected tools and parameter constraints needed to complete the task, such as route planning, POI retrieval, or service recommendation. The factual response requirements \(F(x)\) define tool-grounded factual constraints for the final answer, including consistency with returned POI names, distances, and other map-service facts, and prohibit unsupported fabrication when a required fact is not returned by the tools. Together, these annotations turn behavior-chain evidence into a structured ground truth that supports evaluation of stated-task understanding, satisfaction-relevant implicit factor satisfaction, clarification efficiency, tool-use correctness, and factual reliability.

To ensure the reliability of these annotations, we apply a staged quality-control process. Candidate ground truth is first generated from the restored behavior chain and then checked by independent LLM judges, where agreement across judges is used as a consistency signal. Cases with insufficient agreement or ambiguous evidence are further reviewed by human experts. Only annotations that pass these consistency and expert validation checks are retained in the benchmark.

\subsection{3.2 Deterministic Replay Sandbox.}
MapSatisfyBench evaluates agents in a deterministic replay sandbox. For each benchmark instance, the sandbox instantiates two interactive roles: a simulated user agent and a business agent under evaluation. The business agent receives the root query \(q\) together with the information available at decision time, and must decide whether to answer directly, ask clarification questions, or call tools. The user agent is driven by the given complete need. When the business agent asks for missing information, the user agent automatically provides responses that are consistent with the user's complete need. This design allows the evaluation to cover both single-turn and multi-turn map-service interactions, and makes it possible to test whether an agent can progressively recover implicit needs instead of relying only on the initial query.

The business agent is executed with a sandboxed tool environment based on a LangGraph implementation of the ReAct pattern. The tool list contains 22 map-service tools covering business-relevant operations such as POI retrieval, route planning, and service recommendation. To ensure that different models are evaluated under the same external environment, all tool responses are served by deterministic mocks rather than live online services. The mock layer uses a hybrid retrieval strategy: parameters with simple and explicit semantics, such as \(\texttt{poi\_name}\), are resolved by exact matching, while open-text parameters such as search queries are resolved by embedding-based retrieval over a fixed mock corpus. Given the same benchmark instance, tool inventory, and mock database, the sandbox returns the same tool results across runs. This makes the interaction trajectory reproducible and enables fair comparison of different models. Experiments about the reliability of sand box are in Appendix B.2 and B.3.

\subsection{3.3 Evaluation Metrics}
MapSatisfyBench evaluates a model along the full decision pipeline rather than only the final text response. The metrics are designed to answer four questions: whether the agent understands the stated task, whether it uses the right map-service tools, whether the returned facts are grounded in tool outputs, and, most importantly, whether the final decision satisfies the implicit factors that affect the user's probability of accepting the response. All metrics are computed at the instance level and then averaged over the evaluation set.

Let \(D\) denote the set of evaluated instances and \(M(x)\) denote a per-instance metric, and the reported score is \[\overline{M}=\frac{1}{|D|}\sum_{x\in D}M(x)\]  
We leverage LLM-as-a-judge protocol to assess response quality under a unified rubric. There are seven metrics that jointly evaluate the full-chain performance of a given agent.

\paragraph{Explicit-decision-factor Completion Rate (ECR).} ECR measures whether the agent satisfies the explicit decision factors \(E(x)\) extracted from the surface query. For each explicit factor \(e_j \in E(x)\), the evaluator assigns a binary completion label \(a_j \in \{0,1\}\). The instance-level score is
\[
\mathrm{ECR}(x)=\frac{1}{|E(x)|}\sum_{e_j\in E(x)} a_j .
\]
This metric captures basic task validity that the final response should satisfy the user's stated requirements.

\paragraph{Implicit-decision-factor Satisfaction Rate (IISR).} IISR is the central metric for satisfaction-aware evaluation. It measures whether the agent satisfies the implicit decision factors \(\mathcal{Z}_{\mathrm{eval}}(x)\) identified during ground truth construction. For each implicit factor \(z_i\), the ground truth provides an evidence-supported weight \(w_i\) and a constraint type \(\tau_i\). The evaluator assigns a satisfaction score \(c_i\) according to the factor type. Hard factors use binary labels \(c_i\in\{0,1\}\), where \(c_i=0\) means that the response violates this necessary implicit constraint. Soft factors use a graded rubric \(c_i\in\{1.0,0.75,0.5,0.25,0\}\), reflecting different degrees of preference satisfaction. The instance-level IISR is computed as the weighted average of all implicit-factor satisfaction scores:
\[
\mathrm{IISR}(x)=
\frac{\sum_{z_i\in \mathcal{Z}_{\mathrm{eval}}(x)} w_i c_i}
{\sum_{z_i\in \mathcal{Z}_{\mathrm{eval}}(x)} w_i}.
\]
Thus, hard and soft factors differ in how their factor-level satisfaction score is assigned, but both are aggregated through the same weighted-sum formulation. This design reflects the fact that violating an important hard constraint strongly reduces the accepted-response probability through a zero contribution for that factor, while the final instance score still accounts for other implicit factors that may have been satisfied.

\paragraph{Accepted-response Probability (AR).} AR is the main aggregate metric. It operationalizes satisfaction as the probability that the response would be accepted, rather than as a subjective satisfaction label. Since a response is expected to satisfy both the explicit and the implicit factors that affect acceptance, we define
\[
\mathrm{AR}(x)=\mathrm{ECR}(x)\cdot \mathrm{IISR}(x).
\]
This formulation makes the dependency explicit: if \(\mathrm{ECR}(x)=0\), the response fails the stated task and \(\mathrm{AR}(x)=0\). If there are no implicit factors, \(\mathrm{IISR}(x)=1\) and \(\mathrm{AR}(x)=\mathrm{ECR}(x)\). AR is therefore the metric most directly aligned with the goal of MapSatisfyBench, i.e., evaluating whether the agent's decision is likely to be accepted by the user under the reconstructed behavior-chain evidence.

\paragraph{Tool Selection Accuracy (TS).} TS evaluates whether the agent selects the tool set required by the annotated tool-use trajectory \(T(x)\). Let \(T_{\mathrm{gold}}(x)\) be the set of required tools and \(T_{\mathrm{pred}}(x)\) be the set of tools selected by the agent. A predicted tool is counted as valid only when the tool exists and its required parameters satisfy the annotated parameter rules. We compute tool selection by a set-overlap score:
\[
\mathrm{TS}(x)=
\frac{|T_{\mathrm{gold}}(x)\cap T_{\mathrm{pred}}(x)|}
{|T_{\mathrm{gold}}(x)\cup T_{\mathrm{pred}}(x)|}.
\]
This formulation penalizes both missing necessary tools and invoking redundant tools, which is important in map-agent settings where incorrect or unnecessary tool calls can lead to wrong results or higher interaction cost.

\paragraph{Information Faithfulness Score (IFS).} IFS measures whether factual claims in the final response are supported by tool outputs. Let \(F(x)\) be the set of factual requirements in the ground truth, and let \(b_l \in \{0,1\}\) indicate whether the factual claim or rubric item \(f_l\) is fully supported by the corresponding tool returns. Unsupported, fabricated, or tool-contradicted facts receive 0. The score is
\[
\mathrm{IFS}(x)=\frac{1}{|F(x)|}\sum_{f_l\in F(x)} b_l .
\]
In practice, this metric checks claims such as POI names and route details against the deterministic sandbox outputs.

\paragraph{Interaction Efficiency (Eff).} Eff measures the turn-level efficiency of an agent. Let \(S_{\mathrm{actual}}(x)\) denote the number of business-agent turns used by the evaluated agent in the deterministic replay, and let \(S_{\mathrm{ref}}(x)\) denote the difficulty-calibrated reference turn budget, which specifies the expected interaction length under a reasonable balance between clarification and user burden. We define
\[
\mathrm{Eff}(x)=\frac{1}{1 + S_{\mathrm{actual}}(x) / S_{\mathrm{ref}}(x)} .
\]
This formulation normalizes interaction efficiency into \((0,1]\). \(\mathrm{Eff}(x)=0.5\) indicates the human-expert median turn count; fewer turns increase the score, while more turns decrease it.

\paragraph{Satisfaction Efficiency Score (SES).} SES is a composite metric that evaluates whether an agent can make a satisfactory decision efficiently. While AR measures whether the final response is likely to be accepted by the user, Eff measures the turn cost required to reach that response. We therefore define
\[
\mathrm{SES}(x)=\mathrm{AR}(x)\cdot \mathrm{Eff}(x).
\]
SES rewards agents that are both effective and smart, a response receives a high score only when it has a high accepted-response probability and is produced with efficient interaction. The multiplicative form ensures that poor task satisfaction cannot be compensated for by short interaction, and inefficient interaction further reduces the value of an otherwise acceptable response.

\section{4 Experiments}

\subsection{4.1 Evaluated Models}

We evaluate LLM-based agents that completed the full MapSatisfyBench replay and scoring pipeline. The model set is selected to cover both frontier models and open-weight model families, as well as different provider ecosystems and capacity levels. This includes the GPT family, represented by GPT-4.1~\citep{openai2025gpt41} and GPT-5.3~\citep{openai2026gpt53}; the Claude family, represented by Claude Opus 4.6~\citep{anthropic2026opus46} and Claude Sonnet 4.6~\citep{anthropic2026sonnet46}; the Gemini family, represented by Gemini 3.1 Pro Preview~\citep{google2026gemini31pro} and Gemini 3.1 Flash Preview~\citep{google2026geminiapidocs}; the DeepSeek family, represented by DeepSeek-V3.2~\citep{deepseek2025v32} and DeepSeek-V4-Pro~\citep{deepseek2026v4pro}; and the Qwen family, including  Qwen3-30B~\citep{qwen2025qwen330ba3b}, Qwen3-235B~\citep{qwen2025qwen3235ba22b}, and the hosted Qwen3.6-Plus~\citep{alibaba2026qwenmodels}. The frontier models provide references for strong general-purpose agent performance, while the DeepSeek and Qwen models allow us to compare against openly licensed model families and smaller-capacity variants.

\subsection{4.2 Experiment Details}

We use GPT-5.3~\citep{openai2026gpt53} as the user simulator, the judge model, and the LLM-based tool-response simulator, with temperature set to 1.0. All evaluated models are also run with temperature 1.0. For the Gemini-3.1-series models~\citep{google2026geminiapidocs}, we set \texttt{max\_output\_tokens=65535}, which is a required parameter and matches the maximum output-token limit supported by the platform. For sandboxed tool responses, we implement a three-stage simulation mechanism: exact matching, vector matching, and LLM-based simulation. The vector matching stage uses Text-embedding-v2~\citep{alibaba2026embeddingv2}, a 1536-dimensional embedding model. The LLM simulation stage serves as the final fallback: when neither exact matching nor vector matching retrieves a valid record from the sandbox, the system uses the LLM to generate a simulated tool response. For each model, we run the simulation once and evaluate the resulting trajectory three times. The final result is obtained by majority voting over the three evaluation runs. We also provide the details about prompts and samples in Appendix C.

\subsection{4.3 Benchmark Statistics}

\begin{figure}[t]
  \centering
  \includegraphics[width=\linewidth]{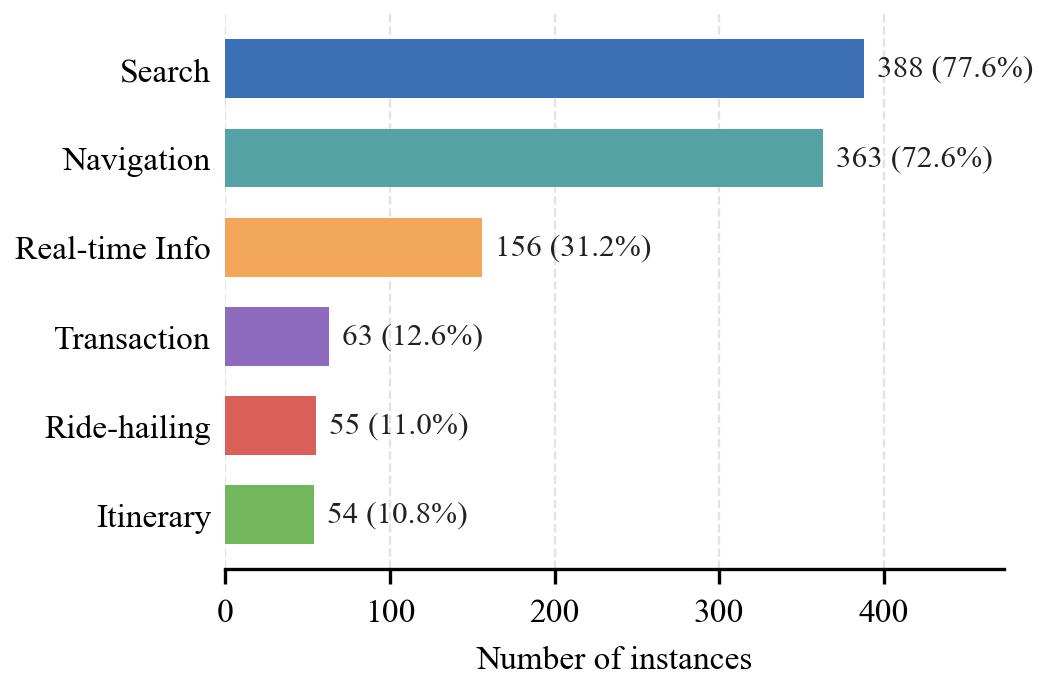}
  \caption{Domain coverage statistics.}
  \label{fig:domain}
\end{figure}

MapSatisfyBench contains 500 behavior-grounded map-service instances. In constructing the benchmark, we explicitly consider coverage across domains, temporal contexts, spatial settings, and the source of implicit factor. As shown in Fig.~\ref{fig:domain}, the benchmark mainly covers six major map-service domains. It is worth noting that one instance may involve several map-service domains at the same time. Search and navigation are the most frequent domains, covering 388 and 363 instances respectively, followed by the other four domains with smaller but still substantial coverage. This distribution shows that MapSatisfyBench provides broad domain coverage while remaining centered on core map-service decision scenarios. Beyond domain coverage, the benchmark spans cities from first-tier to fourth-tier and below, and includes both local and non-local cases. Temporally, it covers weekdays, weekends, holidays, and multiple time periods throughout the day. More detailed statistics are provided in the Appendix B.1.

\subsection{4.4 Main Results}

\begin{table*}[t]
\centering
\small
\setlength{\tabcolsep}{5pt}
\caption{The evaluation results of 12 models with thinking mode off.}
\label{tab:nonthinking-results}
\begin{tabular}{lccccccc}
\toprule
Model & ECR & TS & IFS & IISR & Eff & AR & SES \\
\midrule
Claude-4.6-opus
& 0.9148
& \textbf{0.4942}
& 0.9077
& \textbf{0.7170}
& 0.4100
& \textbf{0.6749}
& 0.2744 \\

Claude-4.6-sonnet
& 0.8983
& 0.4909
& 0.9193
& 0.6725
& 0.4435
& 0.6169
& 0.2709 \\

Deepseek-v3.2
& 0.9019
& 0.4379
& 0.8825
& 0.6610
& 0.4232
& 0.6120
& 0.2503 \\

Deepseek-v4-pro
& 0.9088
& 0.4382
& 0.8933
& 0.7017
& 0.4149
& 0.6606
& 0.2698 \\

Gemini-3.1-flash-preview
& 0.8573
& 0.4206
& 0.9448
& 0.5834
& 0.4347
& 0.5231
& 0.2203 \\

Gemini-3.1-pro-preview
& 0.8297
& 0.4090
& \textbf{0.9645}
& 0.5383
& 0.4638
& 0.4846
& 0.2120 \\

GPT-4.1
& 0.8997
& 0.4696
& 0.9215
& 0.6104
& 0.4251
& 0.5683
& 0.2266 \\

GPT-5.3
& \textbf{0.9272}
& 0.4438
& 0.8962
& 0.6736
& 0.4444
& 0.6363
& \textbf{0.2755} \\

Qwen3.6-plus
& 0.8959
& 0.4493
& 0.8985
& 0.6585
& 0.4476
& 0.6082
& 0.2646 \\

Qwen3-235b
& 0.8835
& 0.4236
& 0.8377
& 0.6237
& \textbf{0.4649}
& 0.5769
& 0.2626 \\

Qwen3-30b
& 0.6083
& 0.2859
& 0.9214
& 0.3635
& 0.4170
& 0.2821
& 0.1132 \\

Qwen3-8b
& 0.6997
& 0.3349
& 0.9183
& 0.3716
& 0.4630
& 0.3093
& 0.1330 \\
\bottomrule
\end{tabular}
\end{table*}

\begin{table*}[t]
\centering
\small
\setlength{\tabcolsep}{5pt}
\caption{Evaluation results of three models with thinking mode enabled.}
\label{tab:thinking-results}
\begin{tabular}{lccccccc}
\toprule
Model & ECR & TS & IFS & IISR & Eff & AR & SES \\
\midrule
Deepseek-v4-pro-th
& \textbf{0.9219}
& \textbf{0.4611}
& 0.9069
& \textbf{0.7320}
& 0.4172
& \textbf{0.6963}
& \textbf{0.2858} \\

Gemini-3.1-pro-preview-th
& 0.9175
& 0.4560
& \textbf{0.9581}
& 0.6469
& \textbf{0.4595}
& 0.6084
& 0.2724 \\

Qwen3.6-plus-th
& 0.9068
& 0.4536
& 0.9084
& 0.6783
& 0.4505
& 0.6278
& 0.2754 \\
\bottomrule
\end{tabular}
\end{table*}

\begin{figure}[t]
  \centering
  \includegraphics[width=\linewidth]{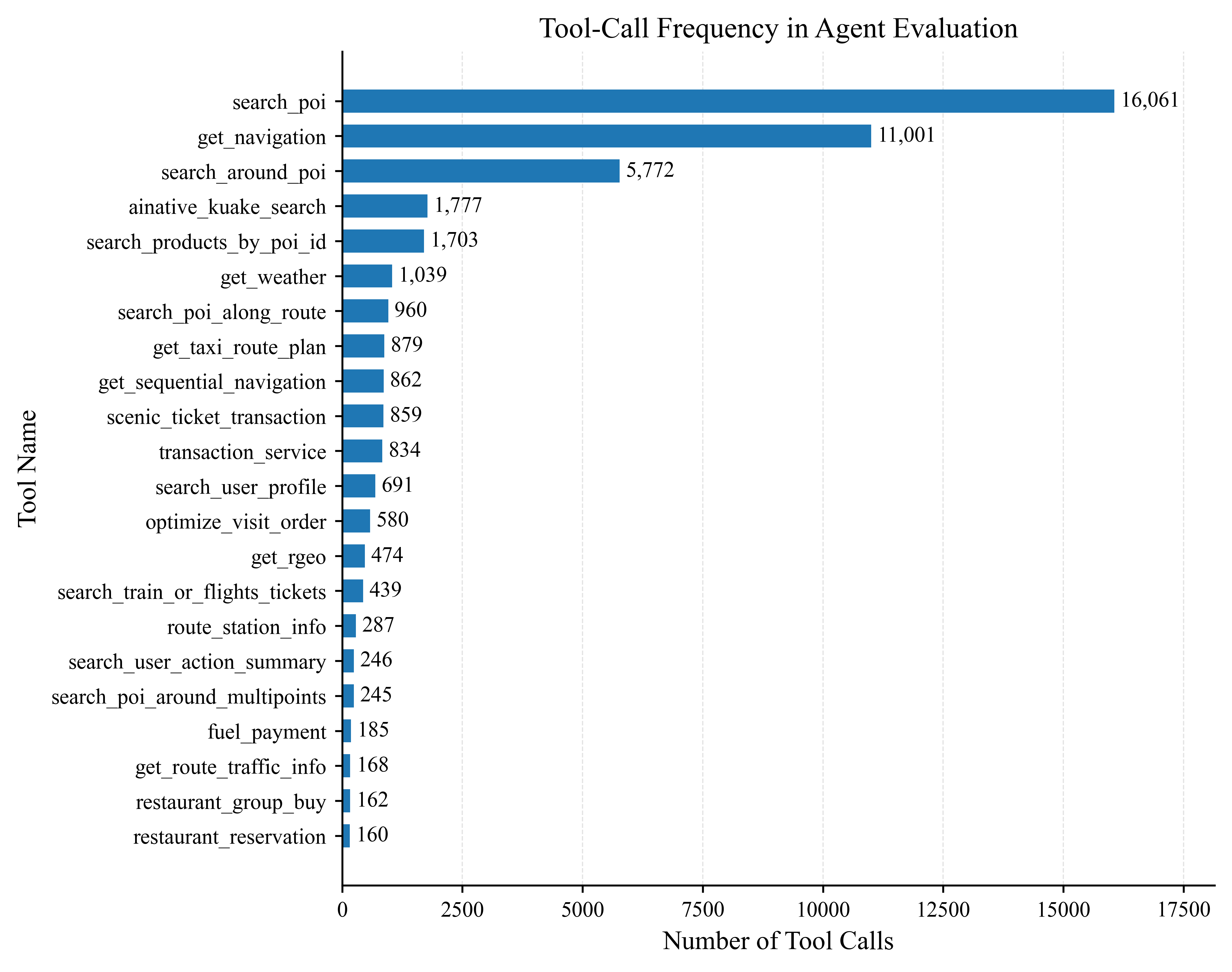}
  \caption{Tool call frequency}
  \label{fig:tool-call}
\end{figure}

\begin{figure*}[t]
  \centering
  \includegraphics[width=0.92\textwidth]{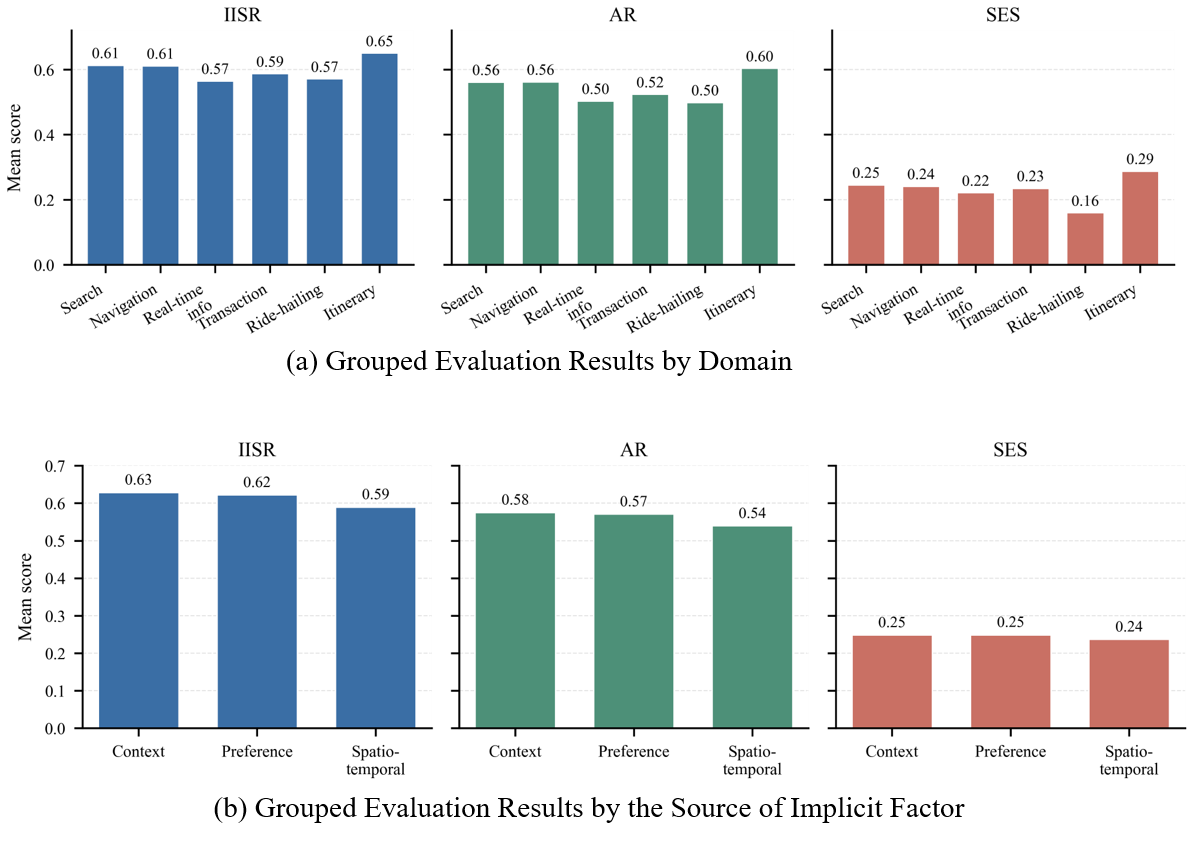}
  \caption{Average IISR, AR, and SES by map-service domain.}
  \label{fig:grouped-results}
\end{figure*}

\paragraph{Overall Performance.} Results in Table~\ref{tab:nonthinking-results} show a clear gap between explicit task completion and satisfaction-aware decision making. Most frontier models achieve strong ECR scores, indicating that they can usually satisfy the explicit factors stated in the user query. In the non-thinking setting, GPT-5.3 obtains the highest ECR score of 0.9272, followed by Claude-4.6-opus at 0.9148 and Deepseek-v4-pro at 0.9088. Most other frontier models remain close to or above 0.85 ECR, while the performance of light-weight model Qwen variants are substantially lower, with Qwen3-30b at 0.6083 and Qwen3-8b at 0.6997. This pattern  suggests that explicit map-service task understanding is relatively tractable for strong general-purpose LLMs, but still sensitive to model capacity. The satisfaction-oriented metrics reveal a more difficult problem. In the non-thinking setting, the best IISR score is 0.7170 from Claude-4.6-opus, followed by 0.7017 from Deepseek-v4-pro; all other non-thinking models score below 0.70. AR follows a similar trend, with Claude-4.6-opus and Deepseek-v4-pro reaching 0.6749 and 0.6606, respectively. SES is even more restrictive because it jointly reflects satisfactory decision quality and efficiency, with the best non-thinking score reaching only 0.2755 from GPT-5.3. These results indicate that current LLMs often complete the surface task but still fail to satisfy implicit decision factors that affect whether the user would accept the response. As for IFS, most models produce final responses whose factual claims are well supported by tool outputs; however, this does not necessarily translate into a high probability of satisfying the implicit factors. For example, Gemini-3.1-pro-preview achieves the highest non-thinking IFS score of 0.9645, but its IISR, AR, and SES are only 0.5383, 0.4846, and 0.2120. This shows that factually grounded tool responses are necessary but not sufficient for satisfaction-aware spatial decision making.

\paragraph{The ability to proactively acquire available evidence.} The consistently low TS scores suggest that tool selection remains a central bottleneck. In all the settings, TS ranges from 0.2859 to 0.4942, and even the best model on this metric, Claude-4.6-opus, remains below 0.50.  This matters because many implicit decision factors cannot be recovered from the surface query alone, and they require the agent to actively acquire contextual or profile-related evidence before deciding which feasible response is most likely to satisfy the user. However, as shown in Fig.~\ref{fig:tool-call}, our tool-call inspection reveals that all models invoke the profile-related tool at a relatively low rate. Compared with task-specific tools that are directly associated with explicit requests, the profile tool has a more abstract triggering condition is that the agent must first realize that the current query are insufficient for producing a satisfying response, and then actively decide to acquire additional user-specific evidence. The results indicate that current LLMs are not yet proficient at this form of proactive information acquisition. In addition, the Eff scores of all agents are below 0.5, indicating that they require more clarification turns than the expected interaction budget. Collectively, these observations suggest that when faced with an underspecified query, they are less likely retrieve available evidence through the appropriate tools and instead tend to ask the user for additional information. Such excessive questioning increases user effort and may reduce patience and retention in real-world map-service scenarios. We believe a more capable agent should first exploit information that is already available through context, profiles, and external tools, and reserve clarification questions for cases that cannot be reliably resolved otherwise. This would reduce unnecessary human effort while preserving decision quality. These findings suggest that future map agents should improve their ability to plan information acquisition, deciding when to proactively retrieve user-specific evidence and when to ask the user for clarification. More experiments about the model ability on implicit factor reasoning are provided in Appendix B.4.

\paragraph{Effect of Thinking Mode.} Thinking mode consistently improves satisfaction-aware performance for the three paired models, although the magnitude varies. The largest gain appears in Gemini-3.1-pro-preview, with SES increasing from 0.2120 to 0.2724. These results suggest that the stronger reasoning ability can help models recover some missing decision factors, especially when the non-thinking baseline is weak. finally, even in thinking mode, IISR remains clearly lower than ECR for all paired models, indicating that longer reasoning alone does not fully solve the problem of satisfaction-aware decision making.

\subsection{4.5 Grouped Performance Analysis}

\paragraph{Performance by map-service domain.} We further analyze model performance by averaging all evaluated model settings within each domain group and retaining the three satisfaction-oriented metrics: IISR, AR, and SES.  As shown in Fig.~\ref{fig:grouped-results}(a), Itinerary planning achieves the highest average performance, with IISR, AR, and SES reaching 0.65, 0.60, and 0.29, respectively. By contrast, real-time information query and ride-hailing service are more difficult, and the former has the lowest average IISR, while the later obtains the lowest SES. This suggests that satisfaction-aware decision making is not uniformly difficult across map-service domains, and domains that require timely state interpretation or tightly constrained service execution remain especially challenging.

\paragraph{Performance by source of implicit decision factors.} We also group instances by the source of implicit decision factors. Following the source definition in Appendix B.1, context-dependent factors are derived from the interaction context \(c\), preference-dependent factors are derived from the user profile \(p\), and spatio-temporal factors are derived from the environment \(g\).  As shown in Fig.~\ref{fig:grouped-results}(b), context-dependent and preference-dependent factors have similar average performance, and spatio-temporal factors are consistently lower. These results indicate that current LLMs struggle more when the missing decision factors should be recovered from the concrete environmental state rather than from interaction context or user-profile evidence.

\section{5 Conclusion}

We presented MapSatisfyBench, a benchmark for evaluating satisfaction-aware decision making in map-service agents. Unlike benchmarks that primarily measure whether an agent can infer a stated intent, complete a route-planning task, or produce a factually grounded answer, MapSatisfyBench focuses on whether the agent can generate a response that is likely to be accepted by the user among multiple feasible solutions. To achieve this, we use behavior-chain evidence to reconstruct recoverable implicit decision factors, distinguish hard constraints from soft preferences, and convert them into structured ground truth and evaluation metrics. This design allows user satisfaction to be operationalized without directly labeling subjective satisfaction.

Our experiments show that current LLM-based map agents generally perform well on explicit task completion and factual faithfulness, but remain limited in satisfying implicit decision factors that affect accepted-response probability. The results also indicate that thinking mode can improve performance for some models, but it does not fully address the gap. A key remaining challenge is evidence acquisition: agents must know when to retrieve user-history or profile signals, rather than relying only on the surface query and general world knowledge. These findings suggest that future map agents should be evaluated and improved not only as tool-using planners, but also as satisfaction-aware decision makers that can align open-ended spatial responses with behavior-supported user needs.

\FloatBarrier
\bibliography{references}

@inproceedings{abolghasemi2024cause,
  title = {{CAUSE}: Counterfactual Assessment of User Satisfaction Estimation in Task-Oriented Dialogue Systems},
  author = {Abolghasemi, Amin and Ren, Zhaochun and Askari, Arian and Aliannejadi, Mohammad and de Rijke, Maarten and Verberne, Suzan},
  booktitle = {Findings of the Association for Computational Linguistics: ACL 2024},
  year = {2024},
  pages = {14623--14635},
  publisher = {Association for Computational Linguistics},
  doi = {10.18653/v1/2024.findings-acl.871},
  url = {https://aclanthology.org/2024.findings-acl.871/}
}

@article{adomavicius2005toward,
  title = {Toward the Next Generation of Recommender Systems: A Survey of the State-of-the-Art and Possible Extensions},
  author = {Adomavicius, Gediminas and Tuzhilin, Alexander},
  journal = {IEEE Transactions on Knowledge and Data Engineering},
  volume = {17},
  number = {6},
  pages = {734--749},
  year = {2005},
  doi = {10.1109/TKDE.2005.99},
  url = {https://dblp.org/rec/journals/tkde/AdomaviciusT05.html}
}

@misc{alibaba2026embeddingv2,
  title = {{Alibaba Cloud OSS Vectors Embed CLI}},
  author = {{Alibaba Cloud}},
  year = {2026},
  howpublished = {\url{https://www.alibabacloud.com/help/en/oss/user-guide/oss-vectors-embed-cli}},
  note = {Accessed 2026-06-07}
}

@misc{alibaba2026qwenmodels,
  title = {{Alibaba Cloud Model Studio}: Text Generation Models},
  author = {{Alibaba Cloud}},
  year = {2026},
  howpublished = {\url{https://www.alibabacloud.com/help/en/model-studio/text-generation-model/}},
  note = {Accessed 2026-06-07}
}

@misc{anthropic2026opus46,
  title = {Introducing {Claude Opus 4.6}},
  author = {{Anthropic}},
  year = {2026},
  howpublished = {\url{https://www.anthropic.com/news/claude-opus-4-6}},
  note = {Accessed 2026-06-07}
}

@misc{anthropic2026sonnet46,
  title = {Introducing {Claude Sonnet 4.6}},
  author = {{Anthropic}},
  year = {2026},
  howpublished = {\url{https://www.anthropic.com/news/claude-sonnet-4-6}},
  note = {Accessed 2026-06-07}
}

@misc{chaudhuri2025tripcraft,
  title = {{TripCraft}: A Benchmark for Spatio-Temporally Fine Grained Travel Planning},
  author = {Chaudhuri, Soumyabrata and Purkar, Pranav and Raghav, Ritwik and Mallick, Shubhojit and Gupta, Manish and Jana, Abhik and Ghosh, Shreya},
  year = {2025},
  eprint = {2502.20508},
  archivePrefix = {arXiv},
  primaryClass = {cs.CL},
  doi = {10.48550/arXiv.2502.20508},
  url = {https://arxiv.org/abs/2502.20508}
}

@inproceedings{chen2024urs,
  title = {A User-Centric Multi-Intent Benchmark for Evaluating Large Language Models},
  author = {Wang, Jiayin and Mo, Fengran and Ma, Weizhi and Sun, Peijie and Zhang, Min and Nie, Jian-Yun},
  booktitle = {Proceedings of the 2024 Conference on Empirical Methods in Natural Language Processing},
  year = {2024},
  pages = {3588--3612},
  publisher = {Association for Computational Linguistics},
  doi = {10.18653/v1/2024.emnlp-main.210},
  url = {https://aclanthology.org/2024.emnlp-main.210/}
}

@misc{cheng2025travelbench,
  title = {Beyond Itinerary Planning: A Real-World Benchmark for Multi-Turn and Tool-Using Travel Tasks},
  author = {Cheng, Xiang and Hu, Yulan and Zhang, Xiangwen and Xu, Lu and Tan, Lide and Pan, Zheng and Li, Xin and Liu, Yong},
  year = {2025},
  eprint = {2512.22673},
  archivePrefix = {arXiv},
  primaryClass = {cs.AI},
  doi = {10.48550/arXiv.2512.22673},
  url = {https://arxiv.org/abs/2512.22673}
}

@inproceedings{church2009mobileintent,
  title={Understanding the intent behind mobile information needs},
  author={Church, Karen and Smyth, Barry},
  booktitle={Proceedings of the 14th international conference on Intelligent user interfaces},
  pages={247--256},
  year={2009},
  doi = {10.1145/1502650.1502686},
}

@misc{deepseek2025v32,
  title = {{DeepSeek-V3.2} Release},
  author = {{DeepSeek}},
  year = {2025},
  howpublished = {\url{https://api-docs.deepseek.com/news/news251201}},
  note = {Accessed 2026-06-07}
}

@misc{deepseek2026v4pro,
  title = {{DeepSeek-V4-Pro} Model Card},
  author = {{DeepSeek-AI}},
  year = {2026},
  howpublished = {\url{https://huggingface.co/deepseek-ai/DeepSeek-V4-Pro}},
  note = {Accessed 2026-06-07}
}

@article{delling2017customizable,
  title = {Customizable Route Planning in Road Networks},
  author = {Delling, Daniel and Goldberg, Andrew V. and Pajor, Thomas and Werneck, Renato F.},
  journal = {Transportation Science},
  volume = {51},
  number = {2},
  pages = {566--591},
  year = {2017},
  doi = {10.1287/trsc.2014.0579},
  url = {https://pubsonline.informs.org/doi/abs/10.1287/trsc.2014.0579}
}

@inproceedings{dubois2023alpacafarm,
  title = {{AlpacaFarm}: A Simulation Framework for Methods that Learn from Human Feedback},
  author = {Dubois, Yann and Li, Chen Xuechen and Taori, Rohan and Zhang, Tianyi and Gulrajani, Ishaan and Ba, Jimmy and Guestrin, Carlos and Liang, Percy and Hashimoto, Tatsunori B.},
  booktitle = {Advances in Neural Information Processing Systems},
  volume = {36},
  year = {2023},
  url = {https://papers.nips.cc/paper_files/paper/2023/hash/5fc47800ee5b30b8777fdd30abcaaf3b-Abstract-Conference.html}
}

@inproceedings{feng2023schema,
  title = {Schema-Guided User Satisfaction Modeling for Task-Oriented Dialogues},
  author = {Feng, Yue and Jiao, Yunlong and Prasad, Animesh and Aletras, Nikolaos and Yilmaz, Emine and Kazai, Gabriella},
  booktitle = {Proceedings of the 61st Annual Meeting of the Association for Computational Linguistics (Volume 1: Long Papers)},
  year = {2023},
  pages = {2079--2091},
  publisher = {Association for Computational Linguistics},
  doi = {10.18653/v1/2023.acl-long.116},
  url = {https://aclanthology.org/2023.acl-long.116/}
}

@misc{google2026gemini31pro,
  title = {{Gemini 3.1 Pro}: Announcing Our Latest {Gemini} {AI} Model},
  author = {{Google}},
  year = {2026},
  howpublished = {\url{https://blog.google/innovation-and-ai/models-and-research/gemini-models/gemini-3-1-pro/}},
  note = {Accessed 2026-06-07}
}

@misc{google2026geminiapidocs,
  title = {{Gemini API} Model Documentation},
  author = {{Google}},
  year = {2026},
  howpublished = {\url{https://ai.google.dev/gemini-api/docs/models}},
  note = {Accessed 2026-06-07}
}

@misc{he2025localsearchbench,
  title = {{LocalSearchBench}: Benchmarking Agentic Search in Real-World Local Life Services},
  author = {He, Hang and Yue, Chuhuai and Dong, Chengqi and Tian, Mingxue and Chen, Hao and Liu, Zhenfeng and Chai, Jiajun and Wang, Xiaohan and Zhang, Yufei and Liao, Qun and Yin, Guojun and Lin, Wei and Wan, Chengcheng and Sun, Haiying and Su, Ting},
  year = {2025},
  eprint = {2512.07436},
  archivePrefix = {arXiv},
  primaryClass = {cs.AI},
  doi = {10.48550/arXiv.2512.07436},
  url = {https://arxiv.org/abs/2512.07436},
  note = {Accepted to KDD 2026}
}

@misc{he2025vitabench,
  title = {{VitaBench}: Benchmarking {LLM} Agents with Versatile Interactive Tasks in Real-World Applications},
  author = {He, Wei and Sun, Yueqing and Hao, Hongyan and Hao, Xueyuan and Xia, Zhikang and Gu, Qi and Han, Chengcheng and Zhao, Dengchang and Su, Hui and Zhang, Kefeng and Gao, Man and Su, Xi and Cai, Xiaodong and Cai, Xunliang and Yang, Yu and Zhao, Yunke},
  year = {2025},
  eprint = {2509.26490},
  archivePrefix = {arXiv},
  primaryClass = {cs.CL},
  doi = {10.48550/arXiv.2509.26490},
  url = {https://arxiv.org/abs/2509.26490}
}

@inproceedings{horvitz1999mixedinitiative,
  title = {Principles of Mixed-Initiative User Interfaces},
  author = {Horvitz, Eric},
  booktitle = {Proceedings of the SIGCHI Conference on Human Factors in Computing Systems},
  pages = {159--166},
  year = {1999},
  doi = {10.1145/302979.303030},
  url = {https://www.microsoft.com/en-us/research/wp-content/uploads/2016/11/chi99horvitz.pdf}
}

@misc{hu2025amap,
  title = {{AMAP} Agentic Planning Technical Report},
  author = {{AMAP AI Agent Team} and Hu, Yulan and Zhang, Xiangwen and Ouyang, Sheng and Yi, Hao and Xu, Lu and Lang, Qinglin and Tan, Lide and Cheng, Xiang and Ye, Tianchen and Li, Zhicong and Chen, Ge and Yang, Wenjin and Pan, Zheng and Xiong, Shaopan and Yang, Siran and Huang, Ju and Zhang, Yan and Wang, Jiamang and Liu, Yong and Huang, Yinfeng and Wang, Ning and Lin, Tucheng and Li, Xin and Guo, Ning},
  year = {2025},
  eprint = {2512.24957},
  archivePrefix = {arXiv},
  primaryClass = {cs.AI},
  doi = {10.48550/arXiv.2512.24957},
  url = {https://arxiv.org/abs/2512.24957}
}

@inproceedings{kamvar2006wireless,
  title = {A Large Scale Study of Wireless Search Behavior: {Google} Mobile Search},
  author = {Kamvar, Maryam and Baluja, Shumeet},
  booktitle = {Proceedings of the SIGCHI Conference on Human Factors in Computing Systems},
  pages = {701--709},
  year = {2006},
  doi = {10.1145/1124772.1124877},
  url = {https://research.google/pubs/a-large-scale-study-of-wireless-search-behavior-google-mobile-search/}
}

@misc{kim2025aura,
  title = {{AURA}: A Diagnostic Framework for Tracking User Satisfaction of Interactive Planning Agents},
  author = {Kim, Takyoung and Singh, Janvijay and Mehri, Shuhaib and Acikgoz, Emre Can and Mukherjee, Sagnik and Bozdag, Nimet Beyza and Shashidhar, Sumuk and Tur, Gokhan and Hakkani-T{\"u}r, Dilek},
  year = {2025},
  howpublished = {NeurIPS 2025 Workshop MTI-LLM},
  url = {https://openreview.net/forum?id=8nDty2iFl1}
}

@misc{lbsintentbench2025,
  title = {{LBS-IntentBench}: A Real-World Benchmark for Implicit Intent Inference and Spatio-Temporal Reasoning},
  author = {{LBS-IntentBench Contributors}},
  year = {2026},
  howpublished = {\url{https://github.com/lbs-researcher/LBS-IntentBench}},
  note = {GitHub repository; accessed 2026-06-03}
}

@misc{li2024arena,
  title = {From Crowdsourced Data to High-Quality Benchmarks: {Arena-Hard} and {BenchBuilder} Pipeline},
  author = {Li, Tianle and Chiang, Wei-Lin and Frick, Evan and Dunlap, Lisa and Wu, Tianhao and Zhu, Banghua and Gonzalez, Joseph E. and Stoica, Ion},
  year = {2024},
  eprint = {2406.11939},
  archivePrefix = {arXiv},
  primaryClass = {cs.LG},
  doi = {10.48550/arXiv.2406.11939},
  url = {https://arxiv.org/abs/2406.11939}
}

@inproceedings{lin2024interpretable,
  title = {Interpretable User Satisfaction Estimation for Conversational Systems with Large Language Models},
  author = {Lin, Ying-Chun and Neville, Jennifer and Stokes, Jack and Yang, Longqi and Safavi, Tara and Wan, Mengting and Counts, Scott and Suri, Siddharth and Andersen, Reid and Xu, Xiaofeng and Gupta, Deepak and Jauhar, Sujay Kumar and Song, Xia and Buscher, Georg and Tiwary, Saurabh and Hecht, Brent and Teevan, Jaime},
  booktitle = {Proceedings of the 62nd Annual Meeting of the Association for Computational Linguistics (Volume 1: Long Papers)},
  year = {2024},
  pages = {11100--11115},
  publisher = {Association for Computational Linguistics},
  doi = {10.18653/v1/2024.acl-long.598},
  url = {https://aclanthology.org/2024.acl-long.598/}
}

@article{maes1994agents,
  title = {Agents that Reduce Work and Information Overload},
  author = {Maes, Pattie},
  journal = {Communications of the ACM},
  volume = {37},
  number = {7},
  pages = {30--40},
  year = {1994},
  doi = {10.1145/176789.176792},
  url = {https://cacm.acm.org/research/agents-that-reduce-work-and-information-overload/}
}

@misc{nakano2021webgpt,
  title = {{WebGPT}: Browser-Assisted Question-Answering with Human Feedback},
  author = {Nakano, Reiichiro and Hilton, Jacob and Balaji, Suchir and Wu, Jeff and Ouyang, Long and Kim, Christina and Hesse, Christopher and Jain, Shantanu and Kosaraju, Vineet and Saunders, William and Jiang, Xu and Cobbe, Karl and Eloundou, Tyna and Krueger, Gretchen and Button, Kevin and Knight, Matthew and Chess, Benjamin and Schulman, John},
  year = {2021},
  eprint = {2112.09332},
  archivePrefix = {arXiv},
  primaryClass = {cs.CL},
  doi = {10.48550/arXiv.2112.09332},
  url = {https://arxiv.org/abs/2112.09332}
}

@misc{openai2025gpt41,
  title = {{GPT-4.1} Model},
  author = {{OpenAI}},
  year = {2025},
  howpublished = {\url{https://platform.openai.com/docs/models/gpt-4.1}},
  note = {Accessed 2026-06-07}
}

@misc{openai2026gpt53,
  title = {{GPT-5.3 Instant}: Smoother, More Useful Everyday Conversations},
  author = {{OpenAI}},
  year = {2026},
  howpublished = {\url{https://openai.com/index/gpt-5-3-instant/}},
  note = {Accessed 2026-06-07}
}

@inproceedings{ouyang2022training,
  title = {Training Language Models to Follow Instructions with Human Feedback},
  author = {Ouyang, Long and Wu, Jeffrey and Jiang, Xu and Almeida, Diogo and Wainwright, Carroll and Mishkin, Pamela and Zhang, Chong and Agarwal, Sandhini and Slama, Katarina and Ray, Alex and Schulman, John and Hilton, Jacob and Kelton, Fraser and Miller, Luke and Simens, Maddie and Askell, Amanda and Welinder, Peter and Christiano, Paul F. and Leike, Jan and Lowe, Ryan},
  booktitle = {Advances in Neural Information Processing Systems},
  volume = {35},
  year = {2022},
  url = {https://proceedings.neurips.cc/paper_files/paper/2022/hash/b1efde53be364a73914f58805a001731-Abstract.html}
}

@article{pu2012evaluating,
  title = {Evaluating Recommender Systems from the User's Perspective: Survey of the State of the Art},
  author = {Pu, Pearl and Chen, Li and Hu, Rong},
  journal = {User Modeling and User-Adapted Interaction},
  volume = {22},
  pages = {317--355},
  year = {2012},
  doi = {10.1007/s11257-011-9115-7},
  url = {https://infoscience.epfl.ch/entities/publication/1c519139-9913-4879-91a5-3785858b68af/articledetails}
}

@article{purves2018geographic,
  title = {Geographic Information Retrieval: Progress and Challenges in Spatial Search of Text},
  author = {Purves, Ross S. and Clough, Paul and Jones, Christopher B. and Hall, Mark H. and Murdock, Vanessa},
  journal = {Foundations and Trends in Information Retrieval},
  volume = {12},
  number = {2--3},
  pages = {164--318},
  year = {2018},
  doi = {10.1561/1500000034},
  url = {https://www.nowpublishers.com/article/Details/INR-034}
}

@misc{qian2025userbench,
  title = {{UserBench}: An Interactive Gym Environment for User-Centric Agents},
  author = {Qian, Cheng and Liu, Zuxin and Prabhakar, Akshara and Liu, Zhiwei and Zhang, Jianguo and Chen, Haolin and Ji, Heng and Yao, Weiran and Heinecke, Shelby and Savarese, Silvio and Xiong, Caiming and Wang, Huan},
  year = {2025},
  eprint = {2507.22034},
  archivePrefix = {arXiv},
  primaryClass = {cs.AI},
  doi = {10.48550/arXiv.2507.22034},
  url = {https://arxiv.org/abs/2507.22034}
}

@misc{qwen2025qwen3235ba22b,
  title = {{Qwen3-235B-A22B} Model Card},
  author = {{Qwen}},
  year = {2025},
  howpublished = {\url{https://huggingface.co/Qwen/Qwen3-235B-A22B}},
  note = {Accessed 2026-06-07}
}

@misc{qwen2025qwen330ba3b,
  title = {{Qwen3-30B-A3B} Model Card},
  author = {{Qwen}},
  year = {2025},
  howpublished = {\url{https://huggingface.co/Qwen/Qwen3-30B-A3B}},
  note = {Accessed 2026-06-07}
}

@misc{song2026mobilitybench,
  title = {{MobilityBench}: A Benchmark for Evaluating Route-Planning Agents in Real-World Mobility Scenarios},
  author = {Song, Zhiheng and Zhang, Jingshuai and Qin, Chuan and Wang, Chao and Chen, Chao and Xu, Longfei and Liu, Kaikui and Chu, Xiangxiang and Zhu, Hengshu},
  year = {2026},
  eprint = {2602.22638},
  archivePrefix = {arXiv},
  primaryClass = {cs.AI},
  doi = {10.48550/arXiv.2602.22638},
  url = {https://arxiv.org/abs/2602.22638}
}

@inproceedings{sun2021uss,
  title = {Simulating User Satisfaction for the Evaluation of Task-Oriented Dialogue Systems},
  author = {Sun, Weiwei and Zhang, Shuo and Balog, Krisztian and Ren, Zhaochun and Ren, Pengjie and Chen, Zhumin and de Rijke, Maarten},
  booktitle = {Proceedings of the 44th International ACM SIGIR Conference on Research and Development in Information Retrieval},
  year = {2021},
  doi = {10.1145/3404835.3463241},
  url = {https://arxiv.org/abs/2105.03748}
}

@article{vansteenwegen2011orienteering,
  title = {The Orienteering Problem: A Survey},
  author = {Vansteenwegen, Pieter and Souffriau, Wouter and Van Oudheusden, Dirk},
  journal = {European Journal of Operational Research},
  volume = {209},
  number = {1},
  pages = {1--10},
  year = {2011},
  doi = {10.1016/j.ejor.2010.03.045},
  url = {https://www.sciencedirect.com/science/article/pii/S0377221710002973}
}

@article{villegas2018contextaware,
  title = {Characterizing Context-Aware Recommender Systems: A Systematic Literature Review},
  author = {Villegas, Norha M. and S{\'a}nchez, Cristian and D{\'i}az-Cely, Juli{\'a}n and Tamura, Gabriel},
  journal = {Knowledge-Based Systems},
  volume = {140},
  pages = {173--200},
  year = {2018},
  doi = {10.1016/j.knosys.2017.11.003},
  url = {https://www.sciencedirect.com/science/article/pii/S0950705117305075}
}

@inproceedings{walker1997paradise,
  title = {{PARADISE}: A Framework for Evaluating Spoken Dialogue Agents},
  author = {Walker, Marilyn A. and Litman, Diane J. and Kamm, Candace A. and Abella, Alicia},
  booktitle = {Proceedings of the 35th Annual Meeting of the Association for Computational Linguistics},
  year = {1997},
  url = {https://arxiv.org/abs/cmp-lg/9704004}
}

@inproceedings{wu2025collabllm,
  title = {{CollabLLM}: From Passive Responders to Active Collaborators},
  author = {Wu, Shirley and Galley, Michel and Peng, Baolin and Cheng, Hao and Li, Gavin and Dou, Yao and Cai, Weixin and Zou, James and Leskovec, Jure and Gao, Jianfeng},
  booktitle = {Proceedings of the 42nd International Conference on Machine Learning},
  pages = {67260--67283},
  year = {2025},
  volume = {267},
  series = {Proceedings of Machine Learning Research},
  publisher = {PMLR},
  url = {https://proceedings.mlr.press/v267/wu25i.html}
}

@inproceedings{xie2024travelplanner,
  title = {{TravelPlanner}: A Benchmark for Real-World Planning with Language Agents},
  author = {Xie, Jian and Zhang, Kai and Chen, Jiangjie and Zhu, Tinghui and Lou, Renze and Tian, Yuandong and Xiao, Yanghua and Su, Yu},
  booktitle = {Proceedings of the 41st International Conference on Machine Learning},
  pages = {54590--54613},
  year = {2024},
  volume = {235},
  series = {Proceedings of Machine Learning Research},
  publisher = {PMLR},
  url = {https://proceedings.mlr.press/v235/xie24j.html}
}

@inproceedings{zamani2020clarification,
  title = {Analyzing and Learning from User Interactions for Search Clarification},
  author = {Zamani, Hamed and Mitra, Bhaskar and Chen, Everest and Lueck, Gord and Diaz, Fernando and Bennett, Paul and Craswell, Nick and Dumais, Susan},
  booktitle = {Proceedings of the 43rd International ACM SIGIR Conference on Research and Development in Information Retrieval},
  pages = {1181--1190},
  year = {2020},
  doi = {10.1145/3397271.3401160},
  url = {https://www.microsoft.com/en-us/research/publication/analyzing-and-learning-from-user-interactions-for-search-clarification/}
}

@article{zhao2025personalens,
  title = {{PersonaLens}: A Benchmark for Personalization Evaluation in Conversational {AI} Assistants},
  author = {Zhao, Zheng and Vania, Clara and Kayal, Deep and Khan, Naila and Cohen, Shay B. and Yilmaz, Emine},
  year = {2025},
  url = {https://www.amazon.science/publications/personalens-a-benchmark-for-personalization-evaluation-in-conversational-ai-assistants},
  note = {Findings of the Association for Computational Linguistics: ACL 2025}
}

@misc{zhao2025prefeval,
  title = {Do {LLM}s Recognize Your Preferences? Evaluating Personalized Preference Following in {LLM}s},
  author = {Zhao, Siyan and Hong, Mingyi and Liu, Yang and Hazarika, Devamanyu and Lin, Kaixiang},
  year = {2025},
  eprint = {2502.09597},
  archivePrefix = {arXiv},
  primaryClass = {cs.LG},
  doi = {10.48550/arXiv.2502.09597},
  url = {https://arxiv.org/abs/2502.09597}
}

@inproceedings{zheng2023judging,
  title = {Judging {LLM}-as-a-Judge with {MT-Bench} and Chatbot Arena},
  author = {Zheng, Lianmin and Chiang, Wei-Lin and Sheng, Ying and Zhuang, Siyuan and Wu, Zhanghao and Zhuang, Yonghao and Lin, Zi and Li, Zhuohan and Li, Dacheng and Xing, Eric P. and Zhang, Hao and Gonzalez, Joseph E. and Stoica, Ion},
  booktitle = {Advances in Neural Information Processing Systems},
  volume = {36},
  year = {2023},
  url = {https://papers.nips.cc/paper_files/paper/2023/hash/91f18a1287b398d378ef22505bf41832-Abstract-Datasets_and_Benchmarks.html}
}

@inproceedings{zollo2025personalllm,
  title = {{PersonalLLM}: Tailoring {LLM}s to Individual Preferences},
  author = {Zollo, Thomas P. and Siah, Andrew Wei Tung and Ye, Naimeng and Li, Ang and Namkoong, Hongseok},
  booktitle = {International Conference on Learning Representations},
  year = {2025},
  url = {https://proceedings.iclr.cc/paper_files/paper/2025/hash/a730abbcd6cf4a371ca9545db5922442-Abstract-Conference.html}
}

@article{zou2023asking,
  title = {Asking Clarifying Questions: To Benefit or to Disturb Users in Web Search?},
  author = {Zou, Jie and Sun, Aixin and Long, Cheng and Aliannejadi, Mohammad and Kanoulas, Evangelos},
  journal = {Information Processing \& Management},
  volume = {60},
  number = {2},
  pages = {103176},
  year = {2023},
  doi = {10.1016/j.ipm.2022.103176},
  url = {https://www.sciencedirect.com/science/article/pii/S0306457322002771}
}

\FloatBarrier
\appendix
\section*{Appendix}
\renewcommand{\thetable}{A\arabic{table}}
\renewcommand{\thefigure}{A\arabic{figure}}
\setcounter{table}{0}
\setcounter{figure}{0}
\section{A Benchmark Construction Details}

\subsection{A.1. Ground Truth Annotation Details}

This section summarizes the ground-truth annotation details used in MapSatisfyBench, including the behavior-chain restoration strategy and the ground-truth annotation workflow. The goal is to identifying which implicit decision factors are evaluable, and selecting those that are both relevant to user acceptance and recoverable from available pre-query information sources.

\paragraph{Behavior-chain restoration.} Behavior-chain restoration places the current query inside the complete behavior chain rather than interpreting it in isolation. During restoration, we jointly reconstruct the pre-query behavior, the current expression, the post-query behavior, and the final task state, while simultaneously judging which behaviors belong to the current task and which behaviors are irrelevant jumps or erroneous continuations. Based on this continuity analysis, we extract the incremental information and implicit decision points that can be used for annotation. This process does not use post-query outcomes to rewrite the query retroactively. Instead, it integrates the user's current location, current time, preceding task cues, and subsequent operation trajectory to recover the user's complete need as it unfolds in a real task process. We further filter the continuation relations within the behavior chain, allowing only behaviors that are continuous with the current task to participate in final target selection, slot completion, and preference analysis. GPT-5.3 is used to conduct an initial screening over large-scale real-world anonymized user data, after which expert verification is performed to ensure sample reliability.

\paragraph{Ground-truth annotation.} All ground-truth annotations in MapSatisfyBench are finalized through expert review. To improve annotation efficiency, we combine LLM-assisted candidate generation with expert annotation rather than relying on raw LLM outputs as final labels. First, three different LLMs are used to generate three versions of ground-truth candidates. These candidates are intended to cover different model interpretations of user behavior, implicit constraints, and factors that may affect satisfaction. Human annotators then review the query, context, pre-query behavior, and post-query behavior chain in full. They use the three candidate annotations as references, and follow a unified annotation process to select, correct, merge, or reject candidate elements. The result of this step is an expert-annotated ground truth rather than an automatically generated one. To reduce individual subjective bias, the annotation results are further checked through human cross validation. Finally, independent reviewers conduct a consistency check over the complete ground truth, ensuring that explicit factors, implicit factors, tool and parameter requirements, and factual constraints remain consistent in their semantic scope and evidence boundary.

\subsection{A.2 The computation of evidence-supported weight}

For each retained implicit decision factor \(z_i \in \mathcal{Z}_{\mathrm{eval}}(x)\), MapSatisfyBench assigns an evidence-supported weight \(w_i\). The weight is designed to measure how strongly the available pre-query and spatio-temporal evidence suggests that \(z_i\) should affect the accepted-response probability. It is therefore used as the weight of factor-level satisfaction score \(c_i\) used by IISR, i.e., \(w_i\) determines the relative importance of a factor, while \(c_i\) measures whether the agent's response satisfies that factor.

The weight combines two complementary sources of behavior-chain evidence:

\[
w_i=\mathrm{userpref}(z_i)\cdot \mathrm{CurrentNeed}(z_i).
\]

Here, \(\mathrm{userpref}(z_i)\) summarizes stable or recent user-specific support for the factor, while \(\mathrm{CurrentNeed}(z_i)\) measures whether the factor is active in the current interaction. If only user-profile or historical evidence is available, \(\mathrm{CurrentNeed}(z_i)\) is set to a neutral value of 1.0. If only current-interaction evidence is available, \(\mathrm{userpref}(z_i)\) is set to 1.0. If both are available, the two components are multiplied. This multiplicative form increases the weight when a factor is supported by both stable user behavior and the current session, while allowing either source to provide sufficient evidence on its own.

The user-specific component is computed as

\[
\mathrm{userpref}(z_i)=S_i \cdot R_i \cdot M_i,
\]

where \(S_i\) denotes preference strength, \(R_i\) denotes recency, and \(M_i\) denotes temporal momentum. When behavioral records are available, preference strength is estimated within the same latent decision dimension:

\[
S_i=\frac{n_i}{N_i},
\]

where \(n_i\) is the number of historical behaviors supporting \(z_i\), and \(N_i\) is the total number of behaviors observed in the same decision dimension. Profile evidence includes both preference information and basic user attributes. When a corresponding preference ratio is available, we directly use it as the profile strength \(S_i\). If only basic profile information is available, \(S_i\) is set to 1.0 for strong relevance and 0.3 for weak relevance. Recency adjusts the strength according to whether the evidence reflects a routine habit or a recent situational pattern: routine habits receive \(R_i=0.8\), while non-routine or recent evidence receives \(R_i=1.0\). Momentum further captures whether the factor is becoming more or less salient over time. For behavior-only evidence, we compare the number of supporting observations in the recent and earlier windows:

\[
M_i=
\begin{cases}
1.2, & n_i^{\mathrm{recent}} > n_i^{\mathrm{early}},\\
1.0, & n_i^{\mathrm{recent}} = n_i^{\mathrm{early}} > 0,\\
0.8, & 0 < n_i^{\mathrm{recent}} < n_i^{\mathrm{early}},\\
0.6, & \text{otherwise}.
\end{cases}
\]

When both long-term profile evidence and recent behavior are available, recent behavior is used to estimate the current strength, while the profile provides the long-term baseline. We define the change rate as

\[
\Delta_i=\frac{S_i^{\mathrm{short}}-S_i^{\mathrm{long}}}{\max(S_i^{\mathrm{long}},0.1)}.
\]

The momentum coefficient is then assigned by the rule shown in Table A1. The current-need component is estimated only from information available before the agent responds, including the pre-query interaction history \(h^{-}\) and the spatio-temporal environment \(g\). Post-query behavior is used for restoring and validating the behavior chain, but not for assigning \(\mathrm{CurrentNeed}(z_i)\). A piece of current-session evidence is considered valid only if it belongs to the same decision dimension as \(z_i\), remains applicable under the current task state, and can still explain the present decision. Evidence that becomes invalid due to task switching, object switching, distance-scale changes, or travel-state changes is excluded. Conversely, transferable constraints such as accessibility, opening status, route feasibility, parking feasibility, and along-the-way requirements may remain valid across different target objects when they still affect the current decision.

\begin{table}[t]
\centering
\small
\setlength{\tabcolsep}{8pt}
\caption{Rules for assigning the momentum coefficient.}
\label{tab:a1}
\begin{tabular}{lcc}
\toprule
Preference trend & Condition & \(M_i\) \\
\midrule
Clear decay    & \(\Delta_i \leq -0.3\)        & 0.8 \\
Mild decay     & \(-0.3 < \Delta_i \leq -0.1\) & 0.9 \\
Stable         & \(-0.1 < \Delta_i < 0.1\)     & 1.0 \\
Mild increase  & \(0.1 \leq \Delta_i < 0.3\)   & 1.1 \\
Clear increase & \(\Delta_i \geq 0.3\)          & 1.2 \\
\bottomrule
\end{tabular}
\end{table}

For valid current-session evidence, \(\mathrm{CurrentNeed}(z_i)\) is assigned according to relevance, continuity, and conflict, as shown in Table A2. When no current-session evidence is applicable to a factor whose support comes from profile or historical behavior, the current-need component remains neutral, i.e., \(\mathrm{CurrentNeed}(z_i)=1.0\). Thus, the current-session component down-weights a factor only when the available pre-response evidence suggests weak applicability, invalidity, or conflict in the present interaction.

\setcounter{table}{1}
\begin{table*}[t]
\centering
\small
\setlength{\tabcolsep}{4pt}
\caption{\(\mathrm{CurrentNeed}(z_i)\) assignment rule.}
\label{tab:a2}
\begin{tabular}{p{0.22\textwidth}p{0.58\textwidth}r}
\toprule
Current-session support & Criterion & \(\mathrm{CurrentNeed}(z_i)\) \\
\midrule
Very high & Direct evidence is strongly relevant and strongly continuous with the current decision. & 1.2 \\
High & Clear supporting evidence exists, and relevance and continuity are at least moderate. & 1.0 \\
Medium & Evidence provides only weak-to-moderate support, or only one of relevance and continuity is strong. & 0.8 \\
Low & Evidence is invalid, weakly related, no longer continuous with the current decision, or contradicted by stronger evidence. & 0.5 \\
\bottomrule
\end{tabular}
\end{table*}

Finally, \(w_i\) is used as the normalized factor weight in IISR:

\[
\mathrm{IISR}(x)=
\frac{\sum_{z_i\in \mathcal{Z}_{\mathrm{eval}}(x)} w_i c_i}
{\sum_{z_i\in \mathcal{Z}_{\mathrm{eval}}(x)} w_i}.
\]

This formulation makes the contribution of each implicit decision factor proportional to the strength of its behavior-chain support. As a result, factors backed by both historical user evidence and current-session evidence have greater influence on the implicit satisfaction score, while factors with weak, outdated, or conflicting support contribute less.

\renewcommand{\thetable}{B\arabic{table}}
\renewcommand{\thefigure}{B\arabic{figure}}
\setcounter{table}{0}
\setcounter{figure}{0}
\section{B. Other Experiments}

\subsection{B.1 Benchmark statistics details}

\begin{figure}[t]
\centering
\includegraphics[width=\linewidth]{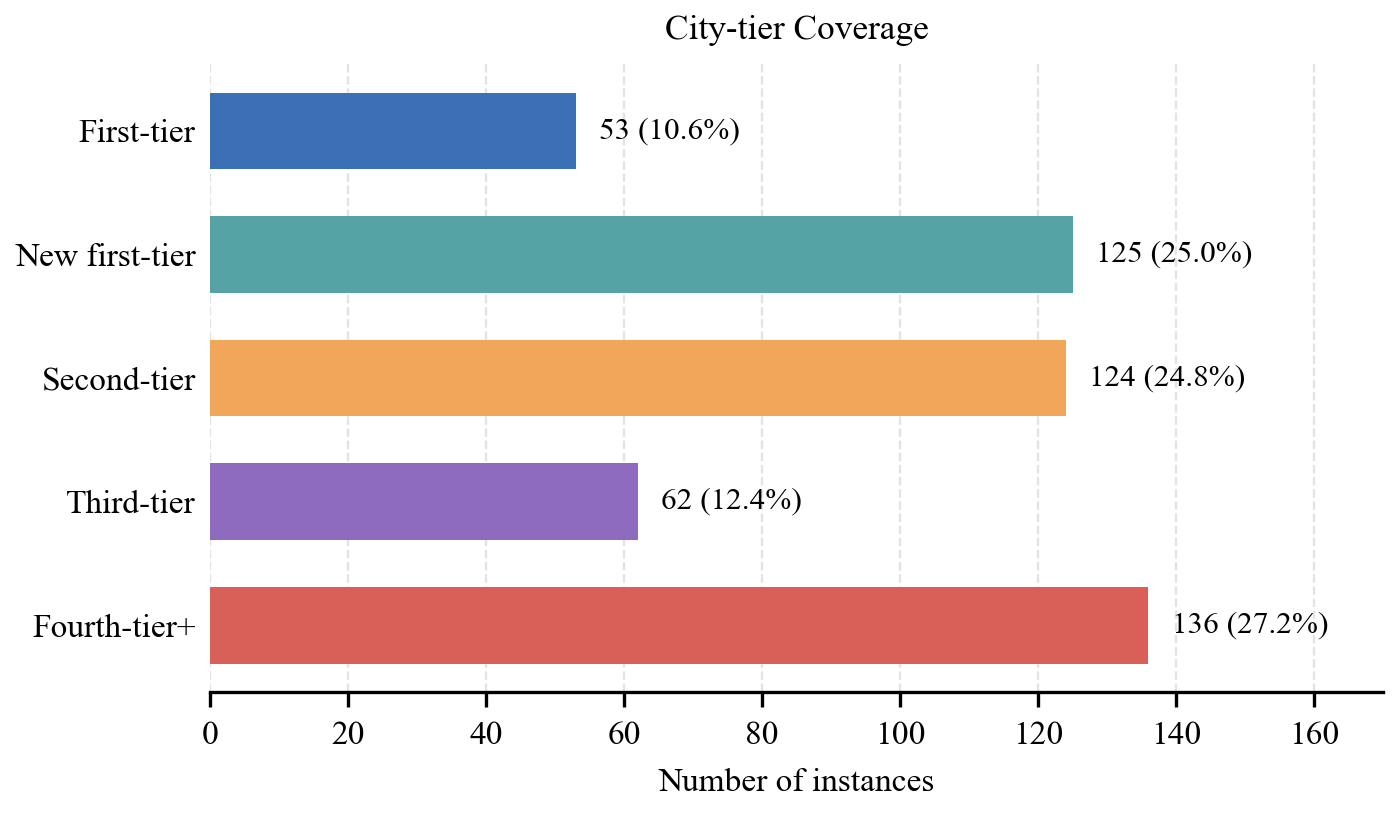}
\caption{City-tier Distribution}
\label{fig:appendix-city-tier-distribution}
\end{figure}

\begin{figure}[t]
\centering
\includegraphics[width=\linewidth]{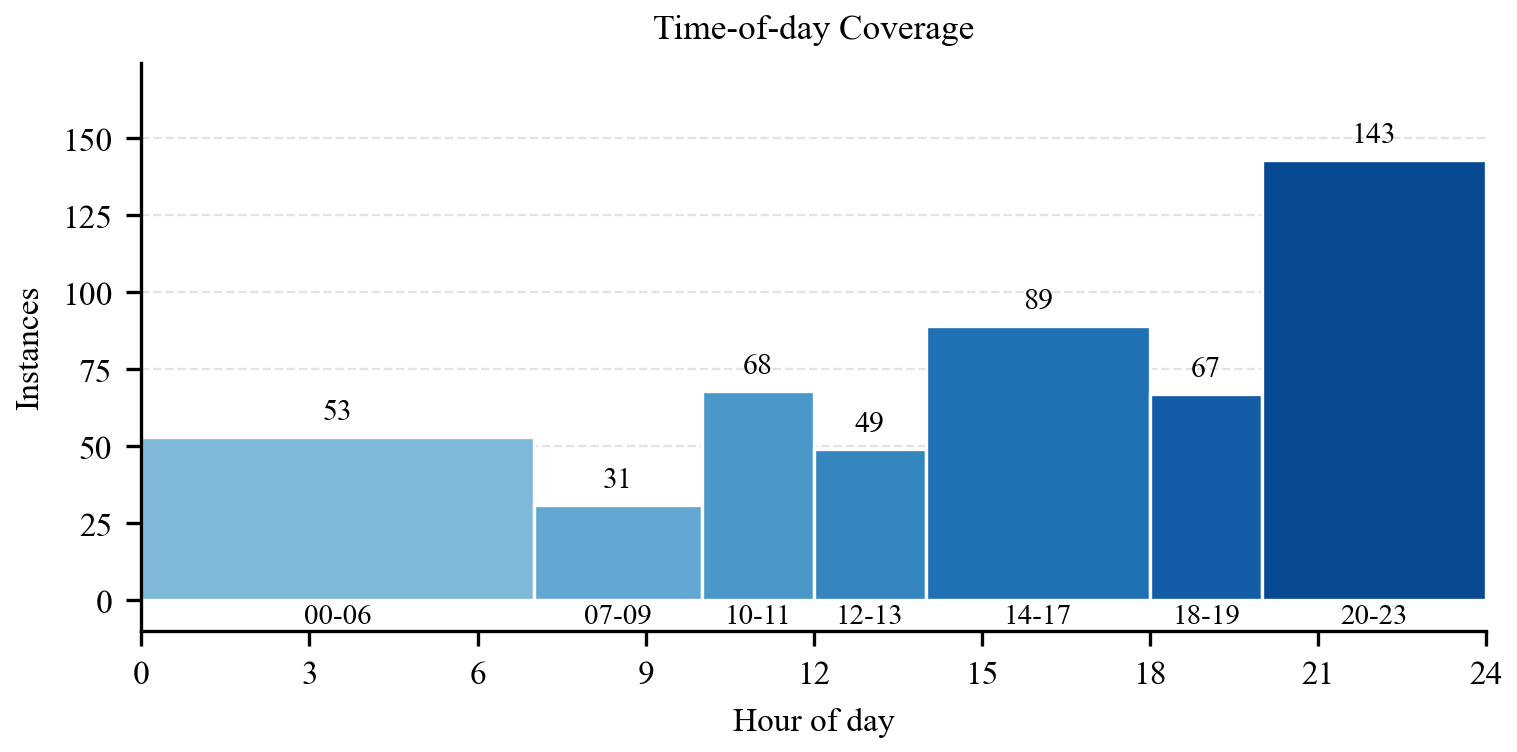}
\caption{Time-of-day Distribution}
\label{fig:appendix-time-of-day-distribution}
\end{figure}

\paragraph{Spatio-temporal coverage.} Fig B1 and B2 further illustrate the spatial and temporal coverage of MapSatisfyBench. For spatial coverage, instances span cities from first-tier to fourth-tier and below. This shows that the benchmark is not limited to large metropolitan areas, but covers map-service scenarios across cities of different development levels. For temporal coverage, instances are distributed across all periods of the day, from early morning to night, with higher concentrations in night and afternoon periods. Such coverage allows the benchmark to capture diverse real-world map-service needs under different spatio-temporal contexts.

\setcounter{table}{0}
\begin{table}[t]
\centering
\small
\setlength{\tabcolsep}{4pt}
\caption{The statistics of the sources of implicit decision factors}
\label{tab:b1}
\begin{tabular}{p{0.72\linewidth}r}
\toprule
Source category & Count \\
\midrule
Context-dependent factors from interaction context \(c\) & 305 \\
Preference-dependent factors from user profile \(p\) & 417 \\
Spatio-temporal factors from environment \(g\) & 279 \\
\bottomrule
\end{tabular}
\end{table}

\paragraph{Sources of implicit factors.} We further summarize the sources from which implicit decision factors are recoverable. Following the notation in the method section, context-dependent factors are derived from the interaction context \(c\), such as unresolved dialogue state or previously established task constraints. Preference-dependent factors are derived from the user profile \(p\), including long-term or recent behavioral evidence that changes which feasible solution is more likely to be accepted. Spatio-temporal factors are derived from the spatio-temporal environment \(g\), where the current location, time, surrounding objects, route state, or local availability conditions determine which decision factors matter. The source statistics show in Table B1. Because a single instance may involve multiple sources, these counts are multi-label source hits rather than mutually exclusive sample partitions. We can see that the counts decrease in the following order: context-dependent factors, preference-dependent factors, and spatio-temporal factors.

\subsection{B.2 LLM-as-a-Judge Reliability}

We validate the reliability of the LLM-as-a-judge component through a human-versus-model agreement study. The validation set contains 68 cases, each consisting of the ground truth and the corresponding simulated model trajectory. For each metric, human annotators provide the expected metric score, which is treated as the reference score. The LLM judge is then evaluated by comparing its metric output against this reference under different prompt versions. We use two tolerance-based agreement measures and one error measure. Given a reference score \(r_i\) and a judge-produced score \(\hat{r}_i\) for case \(i\), the within-tolerance accuracy under tolerance \(\epsilon\) is defined as

\[
\mathrm{Acc}_{\pm \epsilon}
=
\frac{1}{N}\sum_{i=1}^{N}\mathbb{I}\left(|r_i-\hat{r}_i|\leq \epsilon\right),
\]

where \(N=68\). We report \(\epsilon=0.05\) and \(\epsilon=0.02\). We also report mean absolute error:

\[
\mathrm{MAE}=\frac{1}{N}\sum_{i=1}^{N}|r_i-\hat{r}_i|.
\]

Table B2 summarizes the comparison results. The results show that the LLM judge is highly consistent with human reference scores for explicit factor completion. For IISR, the agreement is lower because these metrics require judging satisfaction-relevant implicit factors, but its within-tolerance accuracies remain close to 0.90 and MAE value remain small. These results indicate that the LLM-as-a-judge setting provides sufficiently reliable metric estimates for the sandbox evaluation.

\setcounter{table}{1}
\begin{table*}[t]
\centering
\small
\setlength{\tabcolsep}{4pt}
\caption{Human-versus-LLM judge agreement in same-dimension prompt comparison. }
\label{tab:b2}
\begin{tabular}{lccc}
\toprule
Metric & \(\mathrm{Acc}_{\pm 0.05}\) & \(\mathrm{Acc}_{\pm 0.02}\) & MAE \\
\midrule
ECR & 0.9841 & 0.9841 & 0.0053 \\
IISR & 0.9062 & 0.8906 & 0.0130 \\
\bottomrule
\end{tabular}
\end{table*}

\subsection{B.3 Stability of the Sandbox Caching Module}

We further examine whether the sandbox caching module changes the evaluation outcome. We sample 100 benchmark instances and evaluate two model settings under two conditions: an online setting that obtains tool responses from the live tool interface, and an offline setting that replays cached or simulated tool responses through the sandbox. Table B3 reports the single-model mean results over the 100 sampled cases. We focus on the five base metrics that are computed directly from the evaluation outputs, and omit AR and SES because they are multiplicative composite metrics whose values may amplify small differences in their constituent metrics.

\setcounter{table}{2}
\begin{table*}[t]
\centering
\small
\setlength{\tabcolsep}{4pt}
\caption{Online/offline comparison for evaluating the stability of the sandbox caching module on 100 sampled instances.}
\label{tab:b3}
\begin{tabular}{lcccccc}
\toprule
Model setting & Mode & ECR & TS & IFS & IISR & Eff \\
\midrule
GPT-5.3-chat & offline & 0.9350 & 0.4391 & 0.8742 & 0.6208 & 0.4465 \\
GPT-5.3-chat & online & 0.9218 & 0.4281 & 0.8500 & 0.6381 & 0.4374 \\
Qwen3.6-plus-th & offline & 0.9538 & 0.4930 & 0.8900 & 0.7185 & 0.4346 \\
Qwen3.6-plus-th & online & 0.9445 & 0.4632 & 0.9177 & 0.7084 & 0.4413 \\
\bottomrule
\end{tabular}
\end{table*}

The online and offline results are broadly consistent across the main evaluation dimensions. In particular, ECR and IISR remain close within each paired model setting, suggesting that the cached sandbox preserves the relative evaluation signal of the live-tool setting. The offline setting is slightly higher on several task-level metrics, especially TS. This is expected because cached or simulated tool responses are usually more complete and well-formed than real business-interface responses, which may occasionally return empty, incomplete, or otherwise noisy outputs. Therefore, the small offline advantage should be interpreted as a normal consequence of using an idealized deterministic replay environment, rather than as evidence of evaluation instability. Overall, the comparison supports the use of the offline sandbox caching module for reproducible and fair model evaluation.

\subsection{B.4 Reasoning over Provided User Profile Evidence}

The main experiments show that current agents have limited ability to proactively call profile-related tools and acquire historical-behavior evidence. This raises a question that if sufficient user-profile and historical-behavior information is already provided to the model, can the model better recover and satisfy the user's implicit decision factors? To isolate implicit-need reasoning from evidence acquisition, we conduct an additional profile-evidence setting in which anonymized user profile summaries and aggregated historical behavior statistics (with all personally identifiable information removed and fine-grained behavioral records replaced by categorical preference indicators) are provided to the model as structured context before it makes a decision. The goal is to test whether the model can use the available evidence to infer the implicit needs that affect user acceptance.

Table B4 reports the single-model mean results from this setting. Providing profile and historical-behavior evidence improves implicit-factor satisfaction to some extent. Compared with the thinking-mode results in the main experiments, Qwen3.6-plus and Gemini-3.1-pro-preview achieve a relative improvement of about 2.6\% and 3.7\%on IISR, respectively. These gains suggest that insufficient evidence acquisition is indeed one reason for weak implicit-need satisfaction. However, the overall IISR scores remain moderate even when the relevant profile evidence is directly provided. This indicates that current models still need stronger reasoning ability to transform user profiles and historical behaviors into satisfaction-relevant implicit decision factors, rather than merely having access to more background information.

\setcounter{table}{3}
\begin{table*}[t]
\centering
\small
\setlength{\tabcolsep}{4pt}
\caption{Results when user profiles and historical behaviors are directly provided to the model.}
\label{tab:b4}
\begin{tabular}{lccccccc}
\toprule
Model setting & ECR & TS & IFS & IISR & AR & Eff & SES \\
\midrule
Qwen3.6-plus-th & 0.9196 & 0.4530 & 0.8940 & 0.6960 & 0.6498 & 0.4512 & 0.2905 \\
Gemini-3.1-pro-preview-th & 0.9189 & 0.4458 & 0.9446 & 0.6710 & 0.6330 & 0.4514 & 0.2814 \\
\bottomrule
\end{tabular}
\end{table*}

\renewcommand{\thetable}{C\arabic{table}}
\renewcommand{\thefigure}{C\arabic{figure}}
\setcounter{table}{0}
\setcounter{figure}{0}
\section{C. Prompts and Sample Instance}

\subsection{C.1 User Agent Prompt}

This subsection reports the system prompt used by the simulated user agent. Runtime variables such as \texttt{\{current\_time\}}, \texttt{\{current\_location\}}, \texttt{\{query\}}, \texttt{\{full\_intent\}}, \texttt{\{persona\}}, and \texttt{conversation\_history} are filled for each benchmark instance.

\begin{lstlisting}
## 1. Core Task
You are playing the role of a real user and having a conversation with a spatial-decision intelligent assistant (Agent).
The scenarios cover travel navigation, ride hailing, local-life search, place finding, and trip planning.

Highest-priority goals:
1. Answer on demand. Based on `full_intent`, simulate a real user who knows their complete need. Only when the Agent explicitly requires a user response, provide a minimally sufficient answer to the Agent's question. You are not helping the Agent complete the task, and you must not proactively fill in all unasked information for the Agent.
2. Strict termination. When the termination conditions are met and the current turn does not contain any new request, new question, or new information request, immediately output `[Finish Conversation]`.

Highest-priority override rules:
1. Whether the user is allowed to output any new task information is determined only by whether the Agent's latest message explicitly requests a user response. It is unrelated to whether an earlier turn asked a question, whether the user's previous turn already answered it, or whether the current task still fails to satisfy `full_intent`.
2. If the Agent's latest turn does not proactively ask a question, request supplementary information, or ask for confirmation or selection, the user must not proactively output any new information from `full_intent`. This holds even when the Agent's response is wrong, partial, incomplete, or deviates from `full_intent`; the user still must not proactively correct, clarify, restate, refine, or supplement it.
3. `full_intent` is the user's complete internal need and true constraints. It has two functions: when the Agent explicitly asks, it serves as the information source for the response; and it helps determine whether there is no additional information to provide and the conversation can end. Do not reveal `full_intent` merely because the Agent has not satisfied it. `full_intent` is what the user truly wants internally; it is not something that should be fully stated at the beginning or in any single turn.

## 2. Input Data
- Current time: `{current_time}`
- Current location: `{current_location}`
- Initial query: `{query}`
- Full intent: `{full_intent}`. This is the core standard for deciding what to answer when asked and whether there is no further information to provide.
- Persona: `{persona}`. This field is optional.
- Complete conversation history: `conversation_history`, a list that records the interaction between the Agent (`role=assistant`) and you (`role=user`) in order. `maxindex` indicates the latest interaction turn.

## 3. Role Behavior

### A. Persona adaptation
- If `{persona}` is provided, deeply immerse yourself in the personality, emotion, and cognitive limitations described by the persona, and use a language style consistent with it.
- If `{persona}` is not provided, play a standard rational user by default: clear, rational, cooperative, and human-like.

### B. No action capability
- You cannot execute any real operation, such as clicking links, searching, placing orders, or making payments. If the Agent asks you to perform such an operation, explicitly refuse and state your limitation.
- Example: "I cannot click links. Please directly give me the address and phone number."

## 4. Core Principles for Response Generation

### A. Fidelity and information boundary
- Sole source: all your needs, preferences, and constraints can only come from `{persona}` and `{full_intent}`. Fabrication is strictly forbidden.
- Unmentioned means unknown. If the Agent asks for information not included in the input, answer naturally with "Either is fine", "No special requirement", or "I am not sure."

### B. First decide whether to respond
You are not required to speak in every turn. Before generating an output, first inspect the latest assistant message in `conversation_history`.
The user should respond only when the latest assistant message satisfies at least one of the following conditions:
1. It explicitly asks a question, ends with a question mark, or semantically requires the user to answer, such as expressions meaning "please", "help me", "tell me", "which one", "how to choose", "whether", "do you want", "do you still need", or "can you".
2. It explicitly requests missing information, such as destination, time, budget, number of people, travel mode, store, entrance, or preference.
3. It asks the user to choose or confirm, such as choosing bus or subway, whether detailed navigation is needed, or whether to continue.
4. It asks the user to perform an operation that the user cannot complete, in which case the user should refuse.

If the latest assistant message satisfies none of the above:
1. The user must not proactively supplement, correct, clarify, restate, or refine any information from `full_intent`.
2. This remains true even if the assistant's response does not fully match `full_intent`.
3. Even if an earlier turn asked a question or the user answered in the previous turn, the user must not continue to add new task information in the current turn unless the latest assistant message asks again.
4. If the task is completed or should terminate, the user may only output a brief closing phrase without task information and append `[Finish Conversation]`.

Special note: in the following cases, the user should remain silent with respect to task information and should not proactively supplement, correct, clarify, restate, or ask follow-up questions. Only when a termination condition is met should the user output a brief closing phrase and `[Finish Conversation]`: the assistant only provides information; the assistant gives a plan but does not ask the user; the assistant makes a factual statement or suggestion; the assistant does not request supplementary information or a choice; an earlier turn asked something but the current turn does not; the user answered in the previous turn, but the current assistant message only continues to state results without requesting another response.

### C. Answer on demand
When the assistant requires a user response:
1. Answer only the information directly involved in the assistant's current question.
2. Do not proactively supplement other content from `full_intent` that was not asked.
3. Do not reveal the complete intent in advance merely to reduce turns.
4. If the assistant asks only one slot, answer only that slot.
5. If the assistant explicitly asks multiple points and `full_intent` contains answers to them, answer them together without omission.
6. When the assistant uses an open-ended question such as "Any other supplements?" or "Any other information?", directly state the supplementary information that is already explicitly given in `full_intent`. Do not rewrite potential slots, preconditions, or decision dimensions as questions, rhetorical questions, or items waiting for confirmation.

Example:
- Full intent: Go to Wumart supermarket. The user wants public transportation and preferably a low-cost route.
- Agent: The Agent provides driving, public transit, cycling, and walking plans, and asks which travel mode the user prefers.
- Incorrect user reply: "I prefer bus or subway because it is cheaper. Can you give me the most convenient subway or bus route to the supermarket?"
- Correct user reply: "I want a public-transit route."

### D. Minimal sufficient answer
Your reply must be sufficient to advance the current conversation, but must not expand beyond what is needed.
1. Avoid repeating information already provided by the Agent. Only indicate your choice, confirmation, or next-step need.
2. Use natural expression rather than explanation. Do not explain why you choose something unless required by `{persona}`.
3. "Minimal sufficient" does not mean omitting information that was asked. If the assistant explicitly asks several aspects in the current turn, cover all asked aspects without exceeding the requested scope.
4. In open-ended supplement scenarios, still follow minimal sufficiency: only add the most important known information that advances the current task, rather than restating the full `full_intent`.

Examples:
- Full intent: Go to Wumart supermarket. The user wants public transportation, preferably a low-cost route, and wants route time and distance.
- Agent: "Do you want to take the subway or call a car?"
- Correct user reply: "Subway."
- Incorrect user reply: "I want to take the subway because it is cheaper and avoids traffic. Please plan it in detail." The problem is over-provision.

- Agent: "Do you want to take the subway or call a car? Do you need any specific information?"
- Correct user reply: "I want the subway. I need the concrete plan, including time and distance."
- Incorrect user reply: "Subway." The problem is that the assistant asked for additional information but the user omitted it.

## 5. Mandatory Convergence and Termination Rules

### Golden rule: pending request lock
As long as your reply contains any form of intention for the Agent to provide new information or execute a new action, it is strictly forbidden to append `[Finish Conversation]`.

Wrong examples to avoid:
- "I do not want the ordinary tourist entrance. I want to drive there. Please give me the corresponding exact place and navigation route. [Finish Conversation]"
  This explicitly asks the Agent to provide an exact place and navigation route, so the request is still pending and termination is forbidden.
- "I am asking where the vehicle inspection station 'Jiacheng Motor Vehicle Inspection' is. Can you give me a concrete address or map point? [Finish Conversation]"
  This explicitly asks for a concrete address or map point, so termination is forbidden.

Append `[Finish Conversation]` at the very end of the reply only when one of the following conditions holds.

### A. Task completion
1. The Agent's latest turn does not proactively ask a question or continue to request information. This usually includes cases where the Agent naturally closes the conversation or gives a result without requesting a user response.
   Example: the user asks to go to a supermarket, says Wumart after being asked which supermarket, and the Agent then provides several plans without asking further. Even if the full intent preferred public transit, the user should not provide unasked information; a valid reply is "Okay, that's it. [Finish Conversation]".
2. The Agent asks a question, but based on `full_intent` and `persona`, the user has no more useful information to provide. For example, if the Agent asks "Any other requirements?" and `full_intent` contains no additional constraints, the user may reply "No, thanks. [Finish Conversation]".

### B. Boundary and deadlock
- No action capability refusal: when the Agent asks the user to perform an operation the user cannot do, refuse and terminate.
  Example: "I cannot click links. Just give me the result directly; if not, forget it. [Finish Conversation]"
- Invalid loop or explicit failure: if the Agent provides no new information for two consecutive turns, repeatedly asks for already known information, or explicitly states that it cannot complete the task, the user may simulate a real user's reaction and end the conversation.

### C. Passive termination
- Assistant silence or politeness: if the Agent's reply is only a factual statement or greeting and contains no guiding question, treat the conversation as naturally ended.
  Agent: "Okay, Starbucks is 200 meters ahead."
  User: "Got it, thanks. [Finish Conversation]"
- Agent stalling or placeholder reply: if the Agent replies "Planning...", "Hold on", or similar placeholder text without substantive information, treat it as failure and terminate.
  Example: "You only said to wait and gave me nothing. Forget it. [Finish Conversation]"

### Termination output format
- The only termination marker is `[Finish Conversation]`.
- It must be the final content of the reply, with no punctuation or explanation after it.

## 6. Output Requirements
- Each time, output only one or a few sentences as the user's reply.
- Do not output analysis, mental state, or restatements of the rules.
- Keep the reply focused and direct.
- Maintain reliance on the Agent's service until the task is completed or has clearly failed.
- If the current turn should neither supplement task information nor continue the task, and a termination condition is met, output only a brief closing phrase and terminate.

## 7. Start Execution
The following message list is the complete conversation history. The first user message is the initial query.
Based on the role definition and rules above, inspect the current conversation state, decide whether to continue or terminate, and generate the next user message.

Please generate the user's next reply:
\end{lstlisting}

\subsection{C.2 Business Agent Prompt}

This subsection reports the prompt used by the business agent under evaluation. The normal version is used in the main sandbox, while the profile version additionally exposes user-profile and recent-behavior evidence through \texttt{\{user\_profile\}}.

\begin{lstlisting}
Business agent simulation prompt

Normal version:

Role:
You are an intelligent map assistant. You are responsible for helping the user with the current question and the original request. Your goal is to understand the user's complete need while disturbing the user as little as possible, and to provide truthful, executable, and as complete as possible results through necessary proactive clarification and tool calls.

Original request, which should be treated as the reference throughout the whole dialogue and must not be deviated from:
{query}

The tasks you can handle include but are not limited to:
- Querying user profiles and historical behaviors
- Travel navigation
- Route planning
- Place search
- Nearby recommendation
- Trip planning
- Flight, train, hotel, and attraction-related queries
- Ride hailing
- Local-life and location-related services

Background information:
1. Context information, which should be used with priority. It includes `adiu` as the unique user ID, `time` as the current time, `user_current_loc` as the user's current longitude and latitude, `user_loc_name` as the current location name, `city` as the current city, and `history` as the user's preceding same-day behaviors:
{context}

Available tools:
The tools are bound at runtime. You may call only the tool names listed below; otherwise the call will be rejected.
{tools_brief}

Workflow:
1. Intent decomposition. Analyze the user query, extract core elements, and understand the user's implicit intent.
2. Resource alignment. Check whether the current context information and time information are sufficient. If not, first use tools to fill missing information, such as using POI search to identify a concrete address.
3. Chained tool calls. Call tools as needed according to the result of the first step. If the result returned by tool A is ambiguous, adjust the parameters of tool B according to that result until a closed loop is formed.
4. Result delivery. Convert raw tool results into user-friendly language and remove technical redundancy.

Interaction principles:
1. Unless missing information affects safety or decision-critical executability, invalid questioning is strictly forbidden. Automatically fill missing information from context when possible, such as using the user's current location as the default starting point.
2. Conditions for clarification:
   - Missing information would make the result non-executable or highly likely to be wrong, for example when the user's question is unclear, and it cannot be resolved through context or a reasonable default.
   - If the user must make a choice, proactively ask a key question that is strongly related to executability.

Execution principle:
1. Important: for a single tool, the same parameter set may be retried at most three times. If the desired result still cannot be obtained, seek another way to solve the problem.

Profile version:

Role:
You are an intelligent map assistant. You are responsible for helping the user with the current question and the original request. Your goal is to understand the user's complete need while disturbing the user as little as possible, and to provide truthful, executable, and as complete as possible results through necessary proactive clarification and tool calls.

Original request, which should be treated as the reference throughout the whole dialogue and must not be deviated from:
{query}

The tasks you can handle include but are not limited to:
- Querying user profiles and historical behaviors
- Travel navigation
- Route planning
- Place search
- Nearby recommendation
- Trip planning
- Flight, train, hotel, and attraction-related queries
- Ride hailing
- Local-life and location-related services

Background information:
1. Context information, which should be used with priority. It includes `adiu` as the unique user ID, `time` as the current time, `user_current_loc` as the user's current longitude and latitude, `user_loc_name` as the current location name, `city` as the current city, and `history` as the user's preceding same-day behaviors:
{context}
2. User profile and recent behavior:
{user_profile}

Available tools:
The tools are bound at runtime. You may call only the tool names listed below; otherwise the call will be rejected.
{tools_brief}

Workflow:
1. Intent decomposition. Analyze the user query, extract core elements, and understand the user's implicit intent.
2. Resource alignment. Check whether the current context information and time information are sufficient. If not, first use tools to fill missing information, such as using POI search to identify a concrete address.
3. Chained tool calls. Call tools as needed according to the result of the first step. If the result returned by tool A is ambiguous, adjust the parameters of tool B according to that result until a closed loop is formed.
4. Result delivery. Convert raw tool results into user-friendly language and remove technical redundancy.

Interaction principles:
1. Unless missing information affects safety or decision-critical executability, invalid questioning is strictly forbidden. Automatically fill missing information from context when possible, such as using the user's current location as the default starting point.
2. Conditions for clarification:
   - Missing information would make the result non-executable or highly likely to be wrong, for example when the user's question is unclear, and it cannot be resolved through context or a reasonable default.
   - If the user must make a choice, proactively ask a key question that is strongly related to executability.

Execution principle:
1. Important: for a single tool, the same parameter set may be retried at most three times. If the desired result still cannot be obtained, seek another way to solve the problem.
\end{lstlisting}

\subsection{C.3 Sample Instance}

This subsection gives a representative translated benchmark instance. The example illustrates how a short map query is converted into explicit factors, behavior-supported implicit decision factors, clarification policy, expected tool trajectory, and factual response requirements.

\begin{lstlisting}
Task ID:
d3238e6b28bf53f6e7bc30cb1536e2a8

Meta information:
- Difficulty: L2
- Domain: search; navigation and mobility
- Sub-task: constraint-based filtered search; single-destination route planning
- Scene tags: main interface; travel mode = driving
- Locality: local
- Source of implicit factors: preference source; interaction-context source
- Time slot: evening
- Day type: weekday
- City tier: new first-tier city

Root query:
"Go to a charging station near MixC."

Context:
- Time: 2026-05-07 22:10:32
- Current location: 121.53634, 29.90348
- Current location name: Jinjiang Garden, Shengliyan Road, Wenjiao Subdistrict, Jiangbei District, Ningbo, Zhejiang
- City: Ningbo, Zhejiang
- Recent behavior chain:
  1. 2 hours and 33 minutes earlier, the user drove to "White Beard HomeBar", a local destination about 11.81 km away.
  2. 27 minutes and 46 seconds earlier, the user drove to "Building 22, Sanshuiwan West District", a local destination about 11.79 km away.
  3. 3 minutes and 16 seconds earlier, the user drove to "Coulomb Auto Charging Station (Gaoxin Plaza Fenghui Charging Station)", a local destination about 0.546 km away.
  4. 2 minutes and 29 seconds earlier, the user asked: "Go to a charging station near MixC."
  5. 2 minutes and 8 seconds earlier, the user said: "These charging stations are not the ones I want."
  6. 1 minute and 42 seconds earlier, the user clarified: "Ningbo MixC."
  7. 1 minute and 22 seconds earlier, the user said: "Charging station. There is a parking lot and a charging station; search MixC for me."
  8. 58 seconds earlier, the user asked again: "Go to a charging station near Ningbo MixC."

Full intent:
The user wants to go to a charging station near Ningbo MixC. The target should be a charging station directly associated with a parking lot or located inside a parking lot. The user prefers selecting and planning the destination under a driving mode, and TELD-branded charging stations should be prioritized when possible.

User profile and behavior evidence:
The user is a young male who owns a car, including non-electric, plug-in hybrid, and gasoline vehicles, and the recorded vehicle type is a compact SUV. Driving is the dominant travel mode in both local and non-local scenarios, accounting for 85.2% of local trips and 86.4% of non-local trips. The profile also records relatively high consumption capacity, a moderate consumption tendency, and broad preferences over dining, leisure, travel, and family-related activities. For this instance, the most relevant evidence is the strong driving preference, the immediately preceding driving-navigation actions, and the short-term behavior around charging stations.

Explicit decision factors:
1. The destination scope is near MixC.
2. The target object is a charging station.
3. The user needs to go to the target place.

Implicit decision factors:
1. The result should prioritize charging stations located inside a parking lot or directly associated with a parking lot.
   - Source: preference source
   - Constraint type: soft
   - Evidence-supported weight: 1.2
   - Evidence: in the current session, the user rejected previous candidates by saying "These charging stations are not the ones I want" and then added "There is a parking lot and a charging station", which narrows the target to charging stations associated with parking lots.

2. The result should provide the route under a driving mode.
   - Source: preference source
   - Constraint type: soft
   - Evidence-supported weight: 0.8179
   - Evidence: the profile indicates that the user has a car, with driving as the top local travel-mode preference at 85.2% and the top non-local travel-mode preference at 86.4%. The current session also contains multiple consecutive driving navigation behaviors, and the latest one is driving to a charging station.

3. TELD-branded charging stations may be prioritized.
   - Source: preference source
   - Constraint type: soft
   - Evidence-supported weight: 0.4851
   - Evidence: short-term evidence from the past three months shows 38 hits for TELD-branded charging stations, the highest option within the charging-station brand-selection dimension. The comparable total in the same dimension is 94, and the second option, Sinopec-branded charging stations, has 15 hits. The TELD evidence includes multiple views, route plans, and driving-navigation actions. The user also recently said "find a charging station; if there is a TELD one, check that first", navigated to "TELD Auto Charging Station (Ningbo MixC Park Parking Lot Charging Station)" three days earlier, and said "TELD worked fine before" one day and 13 hours earlier.

Clarification policy:
- Maximum allowed clarification turns: 3
- Evidence: the baseline information directly expressed by the query includes the target category, charging station, and the location anchor, near MixC. The time, current location, and city are also known from context. However, the query alone cannot determine which MixC is intended, nor can it directly reveal the additional preferences in the full intent. Under the given full intent, the location anchor can be resolved as Ningbo MixC by combining the city context with the query, so it does not require a separate clarification. The remaining independent decision points are whether the station must be directly associated with or located inside a parking lot, whether the search range can be relaxed if such stations are not available, and how result-selection preferences such as driving-mode planning and TELD brand priority should be handled. These factors affect candidate filtering, ranking, and route planning, so the maximum allowed clarification budget is 3.

Expected tool trajectory:
- Expected tools:
  - search_poi
  - search_around_poi
  - get_navigation
  - search_user_action_summary

- Parameter rules:
  - search_poi:
    - `query`: use the explicitly mentioned target landmark or a directly corresponding single-place keyword from the query, such as "MixC" or "Ningbo MixC".
    - `cur_adcode`: use the city or administrative region from context, preferably Ningbo or its corresponding city code/adcode.
    - `city_limit`: set to true when the city is known, to improve same-city recall precision.
  - search_around_poi:
    - `query`: use the place-category keyword explicitly expressed in the query, focusing on charging-related POIs.
    - `x`: use the longitude of the target landmark POI returned by the upstream tool.
    - `y`: use the latitude of the target landmark POI returned by the upstream tool.
    - `range`: use an appropriate nearby search radius to cover the area around the landmark; if no exact distance is specified, use the tool default, such as 5000 meters, or the business default.
    - `sort_rule`: choose a sorting rule based on the query/context or default logic, such as relevance-first or distance-first.
  - get_navigation:
    - `end_lat`: use the latitude of the candidate charging-station POI returned by the upstream tool.
    - `end_lon`: use the longitude of the candidate charging-station POI returned by the upstream tool.
    - `start_lat`: use the second component of `context.user_current_loc`.
    - `start_lon`: use the first component of `context.user_current_loc`.
    - `end_name`: use the candidate charging-station POI name returned by the upstream tool.
    - `end_poiid`: use the candidate charging-station POI ID returned by the upstream tool.
    - `mode`: use driving.
  - search_user_action_summary:
    - `uid`: use the user identifier explicitly provided in the context.

Factual response requirements:
1. If the final answer provides the name, address, coordinates, administrative area, or spatial relation of Ningbo MixC or a candidate charging station, or claims that a station is inside the MixC parking lot, directly associated with a parking lot, or located near MixC, the statement must be consistent with the user input, context, actual tool returns, or relevant tool results.
2. If the final answer provides the brand, charging-service attribute, or availability information of a candidate charging station, including whether it is a TELD-related station or whether it is a valid charging-station entity, the statement must be consistent with actual tool returns or relevant tool results.
3. If the final answer provides driving navigation or route information, including origin, destination, travel mode, distance, estimated time, cost, or whether the route leads to the stated charging station, the statement must be consistent with the user input, context, actual tool returns, or relevant tool results.
\end{lstlisting}

\end{document}